\documentclass[onecolumn,conference,compsoc]{IEEEtran}
\usepackage{algorithmic} %format of the algorithm
\usepackage{multirow} %multirow for format of table
\usepackage{amsmath}
\usepackage{xcolor}
\usepackage{bm}
\usepackage{setspace}
\usepackage{geometry}
\usepackage{pifont}
\usepackage{utfsym}
\usepackage{amsmath,amssymb,mathrsfs}

\usepackage{graphicx}
\usepackage{amsmath}
\usepackage{amssymb}
\usepackage{booktabs}
\usepackage{bbm}
\usepackage{bbding}
\usepackage{subcaption}
\usepackage{array}
\usepackage{color}
\usepackage{ragged2e}
\usepackage{bbding}

\usepackage{soul}

\usepackage{amsmath}
\usepackage{url}
\usepackage{fancyhdr}

\begin{document}
\title{SATS: Self-Attention Transfer for Continual Semantic Segmentation}

% \linespread{2.0}
% \usepackage{setspace}

\author{
% Xin Liu$^{1}$, Daqiang Zhang$^{2}$ and Jingyu Zhou$^{3}$% <-this % stops a space
Yiqiao Qiu$^{1}$,Yixing Shen$^{1}$,Zhuohao Sun$^{1}$,Yanchong Zheng$^{3}$\\
Xiaobin Chang$^{2}$,Weishi Zheng$^{1}$,Ruixuan Wang$^{1,}$\textsuperscript{\Envelope}\\
\small{1.School of Computer Science and Engineering, Sun Yat-sen Univerisity, Guangzhou, Guangdong, China}\\
\small{2.School of Artificial Intelligence, Sun Yat-sen Univerisity, Zhuhai, Guangdong, China}\\
\small{3.School of Computer Science, Wuhan University, Wuhan, Hubei, China}
}

\maketitle

\pagestyle{plain}
\thispagestyle{plain}
\pagenumbering{arabic}
\cfoot{1}
% \author[label1]{Yiqiao Qiu}
% \author[label1]{Yixing Shen}
% \author[label1]{Zhuohao Sun}
% \author[label3]{Yanchong Zheng}
% \author[label2]{\\Xiaobin Chang}
% \author[label1]{Weishi Zheng}
% \author[label1]{Ruixuan Wang\corref{cor1}}

% \cortext[cor1]{Correspondence E-mail: wangruix5@mail.sysu.edu.cn}
% \cortext[cor2]{The addresses of Yiqiao Qiu, Yixing Shen and Yanchong Zheng are current address.}
% % \\1:Current Address}
        
% \address{}

% \maketitle
% \begin{abstract}
% %% Text of abstract
% \end{abstract}
\begin{abstract}
Continually learning to segment more and more types of image regions is a desired capability for many intelligent systems. However, such continual semantic segmentation 
{exhibits} catastrophic {forgetting issues similar to those of} continual classification learning. {Unlike the} existing knowledge distillation strategies {for alleviating this problem, transferring} a new type of information, {namely}, the relationships between elements (e.g., pixels) within each image {that} can capture both within-class and between-class knowledge, {is proposed in this study}. Such information can be effectively obtained from self-attention maps in a {Transformer-style} segmentation model. Considering that pixels belonging to the same class in each image {typically} share similar visual properties, a class-specific region pooling operator is novelly applied to provide reliable relationship information for knowledge transfer. Extensive evaluations on multiple public benchmarks reveal that the proposed self-attention transfer method can effectively alleviate the catastrophic forgetting issue. {Furthermore, flexible combinations of the proposed method} with widely adopted strategies {considerably} outperform state-of-the-art solutions.
\end{abstract}

\begin{spacing}{1.5}

\section{Introduction}\label{sec:introduction}

Continually learning knowledge is a desired capability for intelligent systems in many application scenarios, {such as} autonomous driving, autonomous stores, and intelligent healthcare. Most studies on continual learning {have focused} on classification tasks~\cite{LwF, iCaRL, UCIR, LwM} {in which} the classifier is continually updated to learn to recognize more and more classes over multiple stages of continual learning. {In addition to} continual classification tasks, continual semantic segmentation of images {have been} investigated~\cite{ILT,MiB,PLOP,SDR} to continually update a segmentation model such that it can learn to segment more and more types of image region. {Similar to} continual classification, continual semantic segmentation also suffers from the catastrophic forgetting issue~\cite{CatastrophicForgetting}, {that is,} the model {rapidly forgets the} knowledge {obtained from previously learned old classes} after learning to segment more classes of image regions, particularly over multiple stages of continual learning.

\begin{figure}[hbt]
%     \centering
%     \includegraphics[width=1\linewidth]{fig/lines2.pdf}
%     \caption{Comparison between SATS and other methods on VOC 10-1}
%     \label{fig:lines2}
    \centering
    \includegraphics[width=0.6\textwidth]{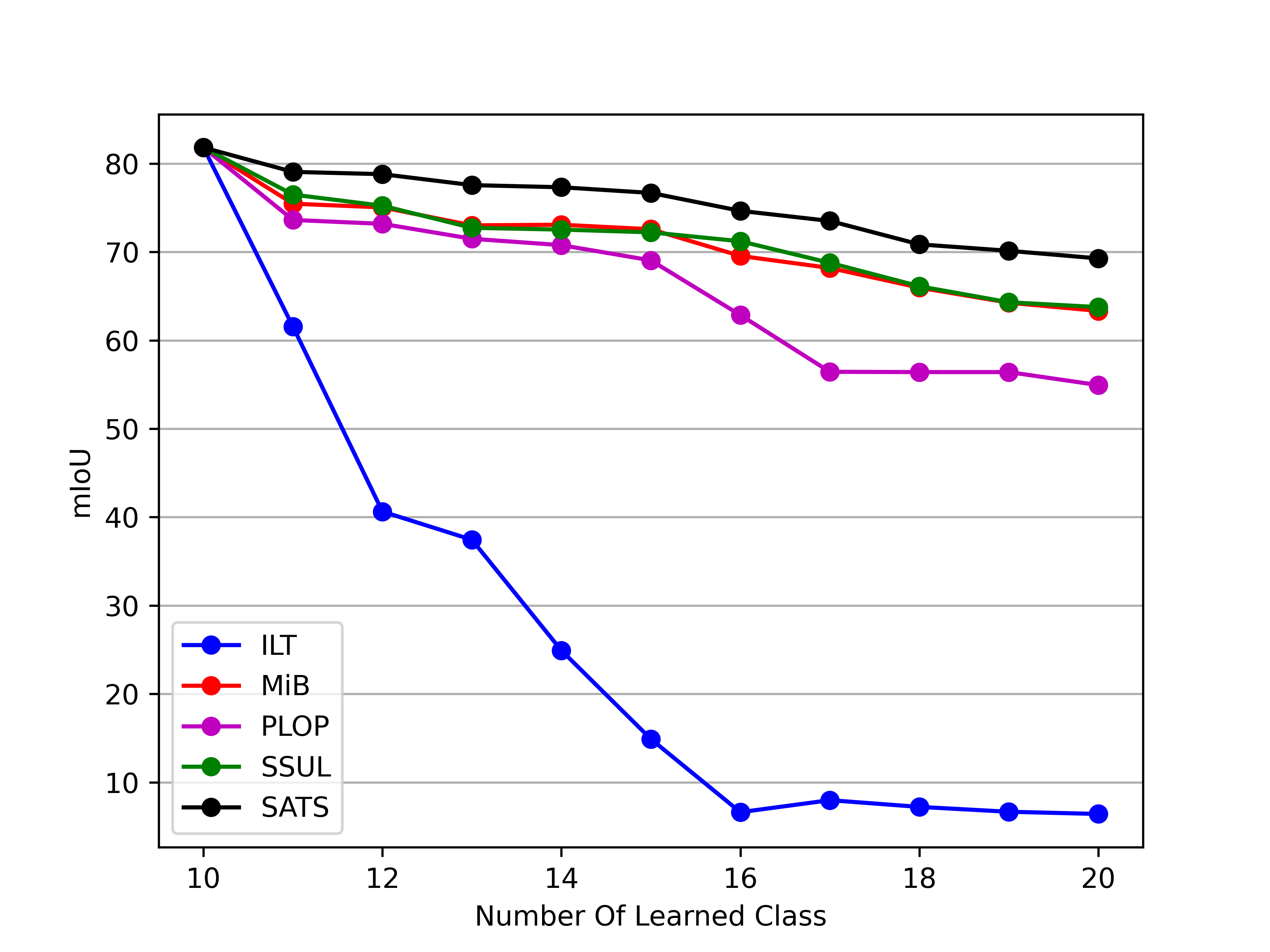}
    \caption{\normalsize Performance {of} the proposed {SATS (Self-Attention Transfer for continual semantic Segmentation) method} and current state-of-the-art methods {in} continual semantic segmentation. Each model with the same backbone initially  {learns} to segment 10 classes of regions and then continually  {learns} to segment one more class at each new stage of continual learning.}
    \label{fig:methodcompare}
\end{figure}

To alleviate the catastrophic forgetting issue, several strategies originally {used} for continual classification tasks have been directly adopted and empirically proved {to be} effective for continual semantic segmentation. One strategy is knowledge distillation, which {is used to} transfer old knowledge by distilling the output of the last layer or multiple layers from the previously updated old model to the current model~\cite{PLOP,PLOPLong}. Another strategy is the {use} of stored exemplars for each previously learned class when training the new model, allowing the new model to directly refresh old knowledge based on these limited old data. {In addition to} directly borrowing ideas from continual classification {tasks}, researchers {have identified} a specific background-shifting issue in continual semantic segmentation tasks, {that is,} background regions in images at one learning stage may contain regions of classes learned at another stage, thus confusing the model when {discriminating} foreground classes against the background class. {Several} effective remedies, including pseudo-labeling of the background regions in new training images based on the previously learned old model~\cite{PLOP, ST-CIL, SSUL} {and} excluding salient regions from the background as {the} unknown (future) class during model learning~\cite{SSUL}{, have been proposed to mitigate this problem}.

In this study, we propose a simple {and} novel knowledge distillation strategy called Self-Attention Transfer for effective continual semantic Segmentation (SATS). {Unlike} existing knowledge distillation methods {in} which {the} model output or visual features {are distilled} from single or multiple convolutional layers, the proposed SATS \textit{distills the visual relationships} between elements (e.g., pixels) within each image {to} capture both within-class and between-class knowledge. Such relational knowledge is obtained from self-attention maps in a transformer-style segmentation model, and it has never been used for continual semantic segmentation in previous studies. {Furthermore,} class-specific region pooling (CRP) is novelly applied to efficiently {elucidate} the visual relationship for both within-class and between-class knowledge. To {the best of} our knowledge, this is the first {study which applies} self-attention and class-region pooling to continual learning. {The proposed SATS can be combined with} widely adopted continual learning strategies {and has achieved} state-of-the-art continual segmentation performance on the VOC~\cite{VOC} and ADE~\cite{ADE20K} semantic segmentation benchmark datasets. The contributions of this study are {as follows}:

\begin{itemize}
    \item A novel knowledge distillation {method is proposed} for continual semantic segmentation. The proposed distillation of relational knowledge is complementary to existing distillation of visual knowledge.
    \item CRP {is applied} to continual semantic segmentation. In particular, CRP is used to extract within-class and between-class knowledge for distillation during continual learning.
    \item This is the first study {in which Transformer is innovatively applied} in continual semantic segmentation.
    \item Extensive evaluations on multiple benchmarks and settings {reveal} that the proposed SATS method can effectively alleviate the catastrophic forgetting issue, and its flexible combinations with widely adopted strategies outperform state-of-the-art methods.
\end{itemize}
% }

\section{Related Work}\label{sec:relatedwork}
% \section{Related Work}
\subsection{Semantic image segmentation}
% {
% In semantic image segmentation, an image is automatically divided 
{Semantic image segmentation task aims to automatically divide an image} into multiple local regions such that each local region corresponds to an object, part of the object, or part of the background. As for other tasks, {such as} image classification and object detection, the state-of-the-art segmentation performance is based on {the} supervised training of a certain deep learning model with a set of labeled training data. Most deep-learning models for semantic segmentation have an encoder–decoder architecture, {in which} the encoder is responsible for extracting features from the input image and the decoder is {used to predict} the semantic label of each image pixel. While fully convolutional networks were previously dominant for both the encoder and decoder, as the segmentation models: FCN~\cite{FCN}, UNet~\cite{UNet}, and the DeepLab series ~\cite{DeeplabV1, DeeplabV2, DeeplabV3, DeeplabV3+}. These CNN-based segmentation models have been outperformed by the transformer-style segmentation models such as SETR~\cite{SETR}, SegFormer~\cite{SegFormer}, and MaskFormer~\cite{MaskFormer}. Such transformer-based segmentation models can learn to extract features for each local region by considering visual proximity between every pair of local regions {using} the self-attention operator at multiple scales in the models.

% {
Besides investigating advanced model backbones for supervised semantic image segmentation, studies {have focused} on challenging conditions, including semi-supervised, weakly supervised, interactive, and domain-adaptive semantic segmentation. {In semi-supervised} semantic segmentation, only a small portion of the training dataset {is assumed to be labeled}. In this case, {the pseudo-labeling} strategy and consistency regularization are {typically} applied to unlabeled training data to train a better segmentation model~\cite{SSSS_1,SSSS_2,SSSS_3}. {Unlike} semi-supervised segmentation, {in} weakly supervised segmentation, each training image is weakly supervised, either in the form of points~\cite{WSSS_point,WSSS_Point2}, scribbles~\cite{WSSS_scribbles2,WSSS_scribbles}, or bounding box~\cite{Lee2021BBAMBB,WSSS_bbox2,WSSS_bbox1}, or image-level labels~\cite{WSSS_RRM, WSSS_L2G, WSSS_OAA}. Depending on the specific form of weak annotations, various strategies, such as the {use of a} class activation map (CAM)~\cite{WSSS_RRM, WSSS_OAA}, have been proposed to estimate more precise regions for objects of interest in the training images~\cite{WSSS_ref5,WSSS_ref6,WSSS_L2G}, and {conventional supervised} learning is applied based on updated and precise annotations. As an extension of the weakly supervised segmentation task, interactive image segmentation allows {humans} to interactively provide feedback to {iteratively} refine the initial segmentation result~\cite{IS_1,IS_2,IS_3_MIDeepSeg}. Such interactive segmentation {is} an efficient strategy {for obtaining} accurate segmentation {results} from the model at the {inference} (i.e., testing) phase and may be practical in real scenarios. Another practical but challenging problem is domain-adaptive semantic segmentation, {in which} the data domain may be {shifted} more or less when applying a trained segmentation model to a {real-world} scenario. In this case, a set of new unlabeled data from the new (target) {domains are typically} collected to fine-tune the {pretrained} segmentation model either {using the} training set from the old (source) domain~\cite{Chen2019DomainAF,You_ref51} or not~\cite{You_MM2021,You_arxiv2021}. {Pseudolabels are typically} estimated for all or part of the unlabeled data, which are then used for model fine-tuning~\cite{You_MM2021}, as in the semi-supervised segmentation. Active learning {can} be applied by estimating a small portion of unlabeled data for humans to annotate~\cite{You_MM2022}, and human-annotated data (together with pseudo-labeled data) are then used to fine-tune the model. Even without any human annotations {or pseudolabels}, the model {can be} fine-tuned during testing, {for example,} by modifying the batch normalization parameters in each model layer with statistics {for the} test data~\cite{You_arxiv2021}. {Techniques} for these challenging segmentation tasks have been {proposed} ( e.g., pseudo-labeling) or could potentially (e.g., active learning) be applied to continual semantic segmentation.

% }

\subsection{Continual learning and continual semantic segmentation}

Most continual learning studies {have focused} on classification tasks~\cite{iCaRL,LwF,UCIR,LwM,End2End-IL}, whereas {limited studies have focused on continual semantic segmentation} using deep learning models~\cite{ILT,MiB,PLOP,SDR,SSUL}. {Generally,} there exist two types of continual learning tasks, task-incremental and class-incremental.
Compared to task-incremental learning which often assumes task identification and the corresponding model head is available during inference, class-incremental learning is more challenging because the model needs to be continually updated to predict all learned classes using a single model head.
%compared two types of continual learning, {namely} task-incremental and class-incremental, {have been proposed with} task-incremental learning in which task identification {is typically assumed,} and  the corresponding model head is available during inference, class-incremental learning is more challenging because the model {should} be continually updated to predict all learned classes {using} a single-model head.
This study {focused} on the class incremental semantic segmentation problem.

% }
% ~\cite{ILT,MiB,PLOP,PLOPLong}
{Numerous approaches} 
developed originally for continual classification may be adapted for continual semantic segmentation.
%developed for continual classification for continual semantic segmentation. 
A widely used {approach of} continual classification is knowledge distillation~\cite{iCaRL,LwF,End2End-IL,LL-PRDR}. {In this method, the} knowledge of previously learned old classes can be demonstrated by the response of the old model to the input data. {Therefore,} expecting that the new model {has} a similar response to the same input would gain or {maintain a} similar old knowledge of the old model. The similarity between the response of the new model and that of the old model can be measured {using the} cross-entropy loss if the response is the model output, as in the LwF~\cite{LwF} and iCarL methods ~\cite{iCaRL}, or generally, by the Euclidean distance or cosine distance when the response is the output of single or multiple intermediate layers, as in the LwM~\cite{LwM} and PODNet methods~\cite{PODNet}. Such knowledge distillation has been effectively applied to continual semantic segmentation, {for example,} by distilling intermediate {features} ~\cite{ILT}, model output~\cite{MiB} or spatially pooled outputs at each intermediate layer~\cite{PLOP,PLOPLong}. {In addition to} knowledge distillation, the replay strategy has {been} proven helpful for continual classification~\cite{iCaRL, IL2M, MCS_rehearsal_continual} and continual semantic segmentation~\cite{SSUL}. A small amount of old data {is stored} for each previously learned class, and the old data {are} combined {with} new classes of training data when updating the model at each subsequent learning stage. Although the stored old data are limited compared with new classes of training data at each learning stage, applying them directly to model update and knowledge distillation can substantially alleviate the catastrophic forgetting of old knowledge~\cite{ES-CIL,AAN-CIL,PEKCL-TRANS}. {Because of} the crucial influence of old data, synthetic old data {have been generated} for continual classification when original data cannot be preserved {because of} privacy or security concerns~\cite{DFCIL}; {this} has been shown {to be} helpful for continual semantic segmentation~\cite{RECALL}.

Since the changing of existing model parameters during continual learning is the {primary} reason of forgetting old knowledge, researchers {have attempted} to keep model parameters relevant to old knowledge from changing during learning new classes of knowledge~\cite{EWC,CIL-DMC}. An example is to fix and combine the major {parts} (feature extractor) of each old model into the new model structure, {and thus expecting to retain} all previously learned old knowledge ~\cite{DER}. {Although such a} model-growing method achieved state-of-the-art performance on continual classification, {maintaining} a balance between model scale and model performance over  {several} stages of continual learning  {remains challenging}~\cite{DER_Bayesian}. In continual semantic segmentation, while state-of-the-art performance was achieved when keeping the segmentation model fixed except for the last layer~\cite{SSUL},
%state-of-the-art  {performance} was achieved  {while} keeping the segmentation model fixed, except for the last layer~\cite{SSUL}. The 
the model is too rigid to learn new classes, and its performance  {degrades rapidly} over more stages. In comparison, our method allows updating of all model parameters for new knowledge learning and provides an (additional) effective  {method} to transfer old knowledge from the old to the new model.
%It is still an open question in effectively balancing model stability and plasticity.

In addition to alleviating the catastrophic forgetting issue, continual semantic segmentation has to solve the background-shifting issue~\cite{MiB,PLOP,SSUL},  {that is, the} background regions in images at one learning stage may  {contain the} regions of the classes learned at another stage. {At each new learning stage,  part of background regions in the new training images can be pseudo-labeled as learned old classes using the old model. Such pseudo-labeling {is} often helpful  for alleviating this issue~\cite{PLOP,PLOPLong,SSUL}}. 
{Besides pseudo-labeling, special consideration of the background class during knowledge distillation from the old model to the new model can help alleviate the background shift issue as well. Specifically, since the old model can predict each image pixel as one of old classes or the background class whereas the new model can additionally predict each image as one of the newly learned classes, the probability of each image pixel belonging to each new classes or the background class predicted by the new model should be aggregated (i.e., summed) and such aggregated probability is then compared to the probability of the pixel belonging to the background class predicted by the old model during knowledge distillation~\cite{MiB}}.
% . This can reduce the background prediction bias
% Aggregating the prediction probabilities of both the background class and new classes from the new class for background class during knowledge distillation can reduce  the bias of background prediction between the old and the new model
{In addition, the detection of} salient regions from image background and considering them as possible future classes (as a separate ‘unknown’ class at the current stage) may reduce background shifting  {in the} future stages of continual learning~\cite{SSUL}. Strategies to alleviate background-shifting issues can be used with those  {for alleviating} catastrophic forgetting in practice. In this study,  {the proposed} method {can also be flexibly combined} with existing strategies for continual semantic segmentation.

 {To date, most} continual semantic segmentation models  {have been} based on fully convolutional network (FCN) backbone.  Besides FCN, the powerful transformer backbone, based on self-attention between elements within each image, has started to exhibit potential ability in solving various computer vision tasks~\cite{PVT, SwinT, Mix-ViT} and Transformer-style segmentation models,  {such as SegFormer~\cite{SegFormer}, have already shown superior} performance than FCN models  {for} semantic segmentation~\cite{DeeplabV3, DeeplabV3+}.  {This study is the first to} evaluate the performance of the Transformer backbone on continual semantic segmentation and use of the unique self-attention information in the transformer to alleviate the catastrophic forgetting issue in continual semantic segmentation.

\section{Method}\label{sec:method}

\begin{figure*}[tp]
    \centering
    \includegraphics[width=\linewidth]{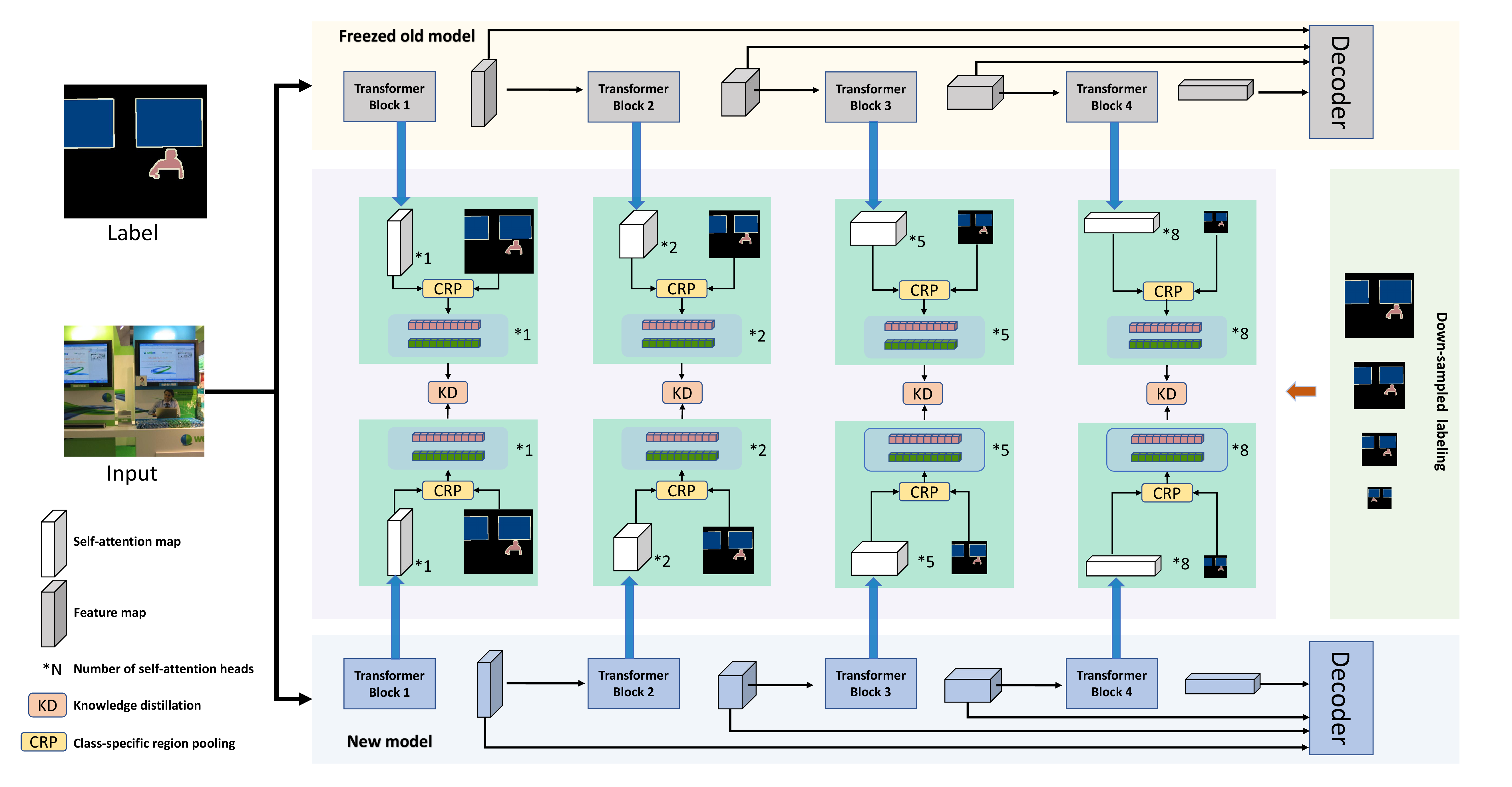}
    \caption{\normalsize Framework of the proposed self-attention transfer method for continual semantic segmentation. Self-attention maps are obtained from the last multi-head self-attention layer in each transformer encoder block, and CRP (also see Figure~\ref{fig:CRP}) is applied to generate a self-attention vector for each foreground class appearing in the input image. The self-attention vectors are then used for within- and between-class knowledge distillation.}
    % and they are reshaped back into 2D for each pixel
    \label{fig:sats}
    
\end{figure*}

The objective of class-incremental semantic segmentation is to continually update a segmentation model  {that} can learn to segment more classes of image regions.  {In} each learning stage, a set of training images  {that correspond} to a specific number of new classes, sometimes together with a small subset of the stored old data for previously learned old classes are used to update the model. Because few or even no old data of previously learned classes are available, the  {primary} challenge of continual semantic segmentation  {is alleviating the rapid} forgetting of old class knowledge by the updated segmentation model, particularly over multiple stages of continual learning. In this study, by  {using} self-attention information from the recently proposed segmentation backbone, a simple yet effective knowledge distillation strategy is proposed to substantially alleviate the catastrophic forgetting issue (Figure~\ref{fig:sats}).

\subsection{Self-attention transfer}

The recently proposed transformer model SegFormer~\cite{SegFormer} achieved state-of-the-art performance on multiple image semantic segmentation tasks~\cite{VOC, ADE20K}. SegFormer and previous FCN-based segmentation models~\cite{UNet,DeeplabV2, DeeplabV3, DeeplabV3+} is the self-attention module at each encoder layer in SegFormer. With self-attention between each element and every other element (corresponding to a single pixel or a small local region) in the input image or feature maps from each SegFormer layer, and the difference between further-apart local regions can be easily learned and globally to help discriminate different classes of pixels or regions. In particular, the elements belonging to the same class will have higher self-attention scores than those from different classes. This  {result leads} to a similar attention-weighted feature vector for the elements that belong to the same classes in the feature map output of the layer. Therefore, self-attention should contain certain essential  {information from} both within-class and between-class information.

Based on  {these observations, we hypothesize} that transferring self-attention information from the old segmentation model to the new model during continual learning may help new model better keep old knowledge.  {This technique differs} from the output feature maps of each layer that represent visual feature information for each input element, self-attention score vectors (from multiple self-attention heads at the layer) for each element contain various relationships between the element and every other element. Thus, self-attention scores within each layer may contain complementary information compared  {with} the feature output of each layer, and distilling such information from the old model to the new model may help the new model remember the knowledge within each old class and across old classes.

\begin{figure}[tbp]
    \centering
    \includegraphics[width=0.7\linewidth]{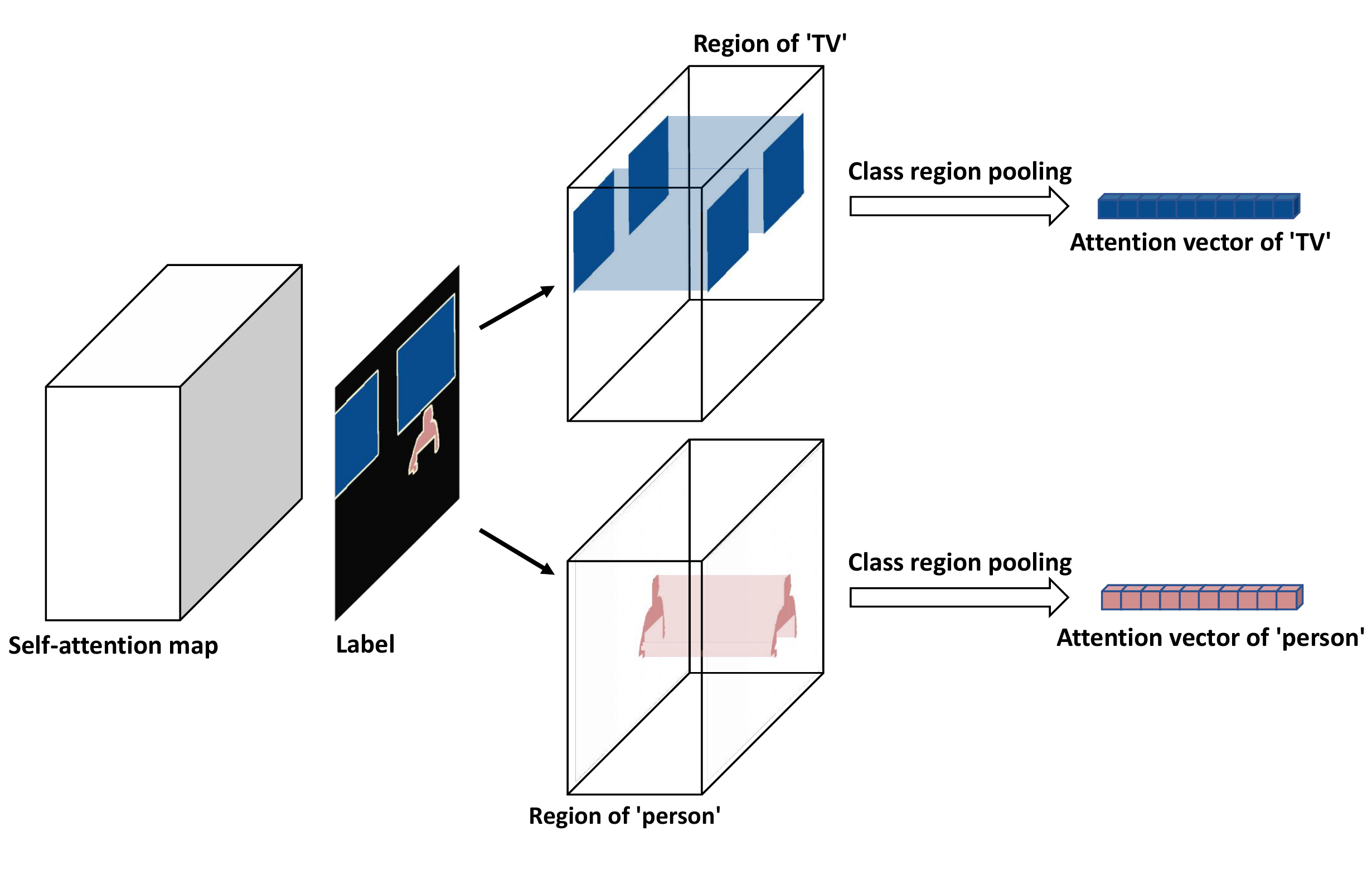}
    \caption{\normalsize Class-specific region pooling.}
    \label{fig:CRP}
\end{figure}

% for each attention layer

% plasticity, 
% robust?
Considering that {distilling} each self-attention score vector at each SegFormer encoder layer  {could prevent} the new model  {from} being flexibly updated to learn new classes of knowledge, we propose applying a class-specific region pooling (CRP) strategy to improve model plasticity during continual learning (Figure~\ref{fig:CRP}). Intuitively, given any specific input image, elements belonging to the same class will have similar self-attention vectors {at each attention layer}. Therefore, pooling self-attention vectors over all elements of the same class results in a single self-attention vector, which is representative  {of} all the elements belonging to the same class. Such pooled self-attention information may even be more robust to small variations or noise in the individual elements.  {Furthermore, because} the feature map output from the last layer of each encoder block is passed to the decoder, the self-attention information from the last self-attention layer in each encoder block contains more information that is useful for segmentation.  {Therefore,} only self-attention information from the last attention layer in each encoder block is considered for distillation (Figure~\ref{fig:sats}), which avoids layer-wise distillation and improves model plasticity during continual learning.

Formally,  we denote $\mathbf{x}_i$ the $i$-th training image, $\mathbf{A}_{i,j,h}$ the three-dimensional self-attention map for the $h$-th attention head at the last attention layer in the $j$-th encoder block, $\mathcal{R}_{i,j,c}$ the set of element locations belonging to the $c$-th class in the layer.  {Subsequently,} the pooled self-attention feature vector for each class from each attention head in each encoder block can be obtained by  {the following equation:}

\begin{equation}
    \mathbf{f}_{i,j,c,h} = \frac{1}{|\mathcal{R}_{i,j,c}|}\sum_{(u,v)\in \mathcal{R}_{i,j,c}}{\mathbf{A}_{i,j,h}(u,v)}
\end{equation}

\noindent where $(u,v)$ represents the location of elements belonging to class $c$. For any specific image, only those classes appearing in the image are considered for the pooling.  {Additionally,} the background region in an image may contain various visual information that is difficult to be represented by one type of  high-level knowledge. Therefore, the background region is excluded  {from} the pooling procedure.  {Here,} $\mathbf{f}^{t-1}_{i,c,h}$ and $\mathbf{f}^t_{i,c,h}$  {denote} the concatenated pooled self-attention vectors over all the encoder blocks from the SegFormer at the previous $(t-1)$-th learning stage and the new SegFormer at the current $t$-th stage, respectively, and $\mathcal{C}_i$ the set of classes (excluding the background class) appearing in image $\mathbf{x}_i$. Next, the distillation of the self-attention information can be achieved by minimizing the loss $L_a(\bm{\theta})$  {as follows,}

\begin{equation} \label{eq:La}
    L_a(\bm{\theta}) = \frac{1}{N\cdot H} \sum_{i=1}^{N} \left\{ \frac{1}{|\mathcal{C}_i|} \sum_{c\in\mathcal{C}_i}  \sum_{h=1}^{H}  || \mathbf{f}_{i,c,h}^t-\mathbf{f}_{i,c,h}^{t-1}||^2 \right\} 
\end{equation}

\noindent where $\bm{\theta}$ is the new model parameters at the $t$-th stage, and $N$ and $H$ represent the number of training images available at the $t$-th stage and the number of attention heads at the last attention layer in the last encoder block,  {respectively.}

% {
 {This study is the} first to apply a class-region pooling (CRP) strategy to continual semantic segmentation.  {Unlike} existing region pooling~\cite{ACFNet, CRP_1, CRP_2}, which  {operate on the} feature map output of the last convolutional layer (representing the class center in the feature space), our CRP operates on self-attention maps (representing relational knowledge within and between classes) of each block of the transformer-style segmentation model. Another difference is in the {role of pooling}. The region pooling in related work~\cite{ACFNet, CRP_1, CRP_2}  is  {typically} part of the segmentation model, and the output of the pooling is used to train a better model under a  {conventional} segmentation setting ( {that is, training} a segmentation model for all classes in a single learning stage).  {By contrast,} pooling in our method is used to distill previously learned knowledge during continual learning.

 {Furthermore,} knowledge distillation in our SATS method  {differs} from existing knowledge distillation for continual learning. Our SATS  {distills} within-class and between-class relational knowledge from self-attention maps during continual semantic segmentation, while existing distillation strategies distill non-relational visual knowledge from the outputs of one or more model layers. Therefore, our method is complementary to existing knowledge distillation methods and can be used to further distill old knowledge during continual learning. Furthermore, knowledge distillation from our SATS is class-specific region-based,  {whereas} existing knowledge distillation is based on the pooling of whole feature maps (as in ILT~\cite{ILT} and MiB~\cite{MiB}) or pooling along various spatial dimensions (as in PLOP~\cite{PLOP}).  In other words, our method tries to respectively distill each class of knowledge from each image, while existing methods distill globally averaged visual information in each image. Considering that multiple categories regions appear in each image, globally averaged visual features are less precise in representing multiple types of knowledge in the input  {rather} than the visual features pooled from each class-specific region.
% }

\subsection{Flexible combination of existing strategies}
The proposed SATS is an independent knowledge distillation strategy and can be easily combined with existing continual learning strategies for semantic segmentation.  {Studies have revealed} that (i) pseudo-labeling of background regions in new classes of images~\cite{PLOP,ST-CIL,SSUL}, (ii) knowledge distillation from the output of the old segmentation model~\cite{MiB, PLOP}, and (iii)  {maintaining} a small number of exemplar images for each old class~\cite{SSUL} keeps model performance in continual learning. Thus, these three strategies were adopted as optional components of continual learning.

 {In the} pseudo-labeling strategy, the old model from the previous learning stage  {is used} to annotate the possible regions of the old classes in the background areas of the new image classes. At each learning stage, only regions of new classes were annotated in the new training images.  {Therefore,} the background regions in the new images may contain regions of previously learned old classes. Annotating and collecting such regions of old classes would  increase the  {amount} of training data for old classes and benefit the training of the new segmentation model.
 %, but also {can avoid from learning incomplete GT (background shifting issue). 
 If the pseudolabels are not available here, the new model will be confused to segment the old classes as background. Following the pseudo-labeling strategy in the PLOP method~\cite{PLOP}, only the background regions with confident predictions as old classes were pseudo-labeled.

For the output distillation strategy, special  {considerations were} made for the background class~\cite{MiB}.  {Because the} image regions of new classes are  {typically} learned as part of background regions by old models in previous learning stages, for each training image at the new learning stage, the MiB method~\cite{MiB} adds the prediction probabilities of the background class and all the new classes from the new model as the modified unbiased prediction of the background class, and such modified prediction is compared to the output of the old model for each input image during knowledge distillation.

As in continual classification tasks, storing a small subset of images for each old class can substantially improve the performance of the new segmentation model for old classes.  {In the} state-of-the-art SSUL method~\cite{MiB}, a memory buffer with a limited size is provided to store old data, and an equivalent number of images for each old class were stored in memory.  {Some of the} stored old data may need to be discarded to store data for more recently learned classes.
% how to discard, memory size: to be mentioned in experimental setting.

It is worth noting these three strategies are supplementary to each other and can be combined with the proposed method. In existing state-of-the-art methods, two or more strategies {are always combined} during continual segmentation learning. For example, in MiB and PLOP, knowledge distillation and pseudo-labeling strategies {are used, whereas in} SSUL, knowledge distillation, pseudo-labeling, and example replay strategies are used. In this study, the pseudo-labeling strategy and knowledge distillation from the model output are combined with the proposed method by default, and the inclusion of the memory buffer to store small old data is optional. Overall, the new segmentation model at each new learning stage is updated by minimizing the combined loss function $L(\bm{\theta})$ over all current new classes of training images at the new learning stage and the stored small old data for each old class {as follows:}

\begin{equation}
    L(\bm{\theta}) = L_{c}(\bm{\theta}) + \lambda_a L_{a}(\bm{\theta}) + \lambda_d L_d(\bm{\theta}) \,
\end{equation}

\noindent where $L_{c}(\bm{\theta})$ is the conventional cross-entropy loss, and $L_a(\bm{\theta})$ is the self-attention transfer loss (Equation~\ref{eq:La}) based on the CRP of the self-attention maps from the old and the new models; here, $L_d(\bm{\theta})$ is the unbiased knowledge distillation loss based on the outputs of both the old and new models~\cite{MiB}, and $\lambda_a$ and $\lambda_d$ are {hyperparameters} to balance the loss terms. 

\section{Experiments}\label{sec:experiments}

\subsection{Settings}
\noindent \textbf{Dataset:}  Following existing work on continual semantic segmentation, we used classical image segmentation benchmark {datasets named} Pascal VOC 2012~\cite{VOC} and ADE20K~\cite{ADE20K} for both quantitative and qualitative evaluations. The VOC dataset includes 20 foreground classes and one background class, with 10582 images for training, 1449 for validation, and 1449 for testing. The ADE20K dataset includes 150 foreground classes and one background class, 20200 images for training, 2000 for validation, and 2000 for testing.

\noindent\textbf{Protocols:} Following the “m-n” protocols used in previous studies, we used the first m foreground classes and the background class in the dataset to train an initial segmentation model. {The model was then} updated over the stages of continual learning by each continual learning method, with each stage learning new classes. We used protocols VOC 15-1, 15-5, 5-3, and 10-1 and ADE 100-10 and 25-25. The segmentation model was initially trained and then updated in six stages (one initial stage plus five continual learning stages), two stages, six stages, and 11 stages for the four protocols on VOC and six stages on ADE. On the ADE20K dataset, the ADE 25-25 {and 100-10 protocols were used because} the 25-25 protocol is more challenging and practical in real applications. {Based on} previous studies, we used the {\textit{overlapped}} setting, {that is,} in each training and validation image at each stage of learning, only the image regions belonging to classes that are learned at the current stage are considered as foreground regions, and other regions are considered as background, even if some of the regions belong to foreground classes that have been learned or will be learned in the future learning stages. Similarly, for test images, only the foreground classes that have been learned at the current or previous stages are considered foreground regions, and all other regions are considered background.
% tense all modify to current tense, modify were to are, had to have 

\noindent\textbf{Metrics:} Standard mean Intersection-over-Union (mIoU) over classes at each learning stage was used to evaluate the performance of each {of these methods.} For detailed investigation, the mIoUs were calculated respectively over the first set of classes in the initial learning stage, and the other classes over subsequent continual learning. stages, and all the classes for each method with each protocol.

\noindent\textbf{Implementation details:} SegFormer-B2~\cite{SegFormer} was used as the default model, whose backbone encoder MixTransformer-B2 was pretrained on ImageNet~\cite{ImageNet} following previous studies~\cite{ILT,MiB,PLOP,SDR,SSUL} on continual semantic segmentation, and the SegFormer decoder was randomly initialized. For the baseline methods {using} Deeplab V3, the model backbone is ResNet-101, {according to} previous studies~\cite{MiB, PLOP}. Also following previous studies~\cite{MiB,PLOP}, during model training and updating of SegFormer-B2 or DeepLab V3, the SGD optimizer with a batch size of 24 for VOC, 32 for ADE, and momentum of 0.9, was used with an initial learning rate of 0.01 for the first (initial) learning stage, 0.001 for subsequent learning stages, and then exponentially decreased with a decay rate of 0.9 over epochs. Each model was trained over 30 epochs and was selected from those with the best performance on the validation set. The hyperparameters $\lambda_a$ and $\lambda_{d}$ were empirically set to 20. For the experiments using memory to store old data, the memory size was set to 100 for VOC and 300 for ADE for all methods following the setting from SSUL~\cite{SSUL}. All experiments were {performed} on PyTorch 1.8 with CUDA 10.2 using two NVIDIA V100 GPUs.

\begin{table*}[t]
\caption{
\normalsize 
Semantic segmentation performance after finishing the last stage of continual learning. With each protocol, model performance on the set of classes learned at the first learning stage (e.g., class 0-15 with protocol VOC 15-1), on the other classes learned over stages of continual learning (e.g., class 16-20 with VOC 15-1), and on all the classes (All) were reported for each method. Methods with * {indicate} the results were directly {obtained} from the corresponding original work, and all the {other results} were based on our re-implementations of these methods. Methods with “-M” denote that the memory of limited old data was used. In each column, the {numbers in bold represent} the highest performance, and the {underlined numbers represent} the second-highest performance.
} % \caption
\centering
\resizebox{\linewidth}{!}{ %< auto-adjusts font size to fill line
\begin{tabular}{ c|c|ccc|ccc|ccc|ccc } 
\toprule
&& \multicolumn{3}{c|}{VOC 15-1 (6 Tasks)} & \multicolumn{3}{c|}{VOC 15-5 (2 Tasks)} & \multicolumn{3}{c|}{VOC 5-3 (6 Tasks)} & \multicolumn{3}{c}{VOC 10-1 (11 Tasks)}\\
Method & Network & 0-15 & 16-20 & All & 0-15 & 16-20 & All & 0-5 & 6-20 & All & 0-10 & 11-20 & All \\ 
\midrule
Joint Training & DeepLab V3 & 79.77 & 72.35 & 77.43 & 79.77 & 72.35 & 77.43 & 76.91 & 77.63 & 77.43 & 78.41 & 76.35 & 77.43\\
%lwf-mc ilt Backbone 存疑
LwF-MC$^{\star}$ \text{\cite{iCaRL}} & DeepLab V3 & 6.40 & 8.90 & 6.90 & 58.10 & 35.00 & 52.30 & 20.91 & 36.67 & 24.66 & 4.65 & 5.90 & 4.95 \\
ILT$^{\star}$ \cite{ILT} & DeepLab V3 & 8.75 & 7.99 & 8.56 & 67.08 & 39.23 & 60.45 & 22.51 & 31.66 & 29.04 & 7.15 & 3.67 & 5.50 \\
MiB$^{\star}$ \cite{MiB} & DeepLab V3 & 35.1 & 13.5 & 29.7 & 75.5 & 49.4 & 69.0 & 57.1 & 42.5 & 46.7 & 12.2 & 13.1 & 12.6 \\
PLOP$^{\star}$ \cite{PLOP} & DeepLab V3 & 66.25 & 24.86 & 56.4 & 75.49 & 49.66 & 69.34 & 17.48 & 19.16 & 18.68 & 44.03 & 15.51 & 30.45 \\ 
SDR$^{\star}$  \cite{SDR} & DeepLab V3+ & 44.7 & 21.8 & 39.2 & 75.4 & 52.6 & 69.9 & N/A & N/A & N/A & N/A & N/A & N/A \\
% SSUL$^{\star}$ & Deeplab V3 & 78.06 & 28.54 & 66.27 & 71.42 & 47.16 & 70.21 & 71.17 & 45.38 & 52.75 & 73.78 & 41.13 & 58.23 \\ 不需要加入ssul没有m的了
RECALL$^{\star}$ \cite{RECALL} & DeepLab V2 & 65.7 & 47.8 & 62.7 & 66.6 & 50.9 & 64.0 & N/A & N/A & N/A & 59.5 & 46.7 & 54.8\\
ST-CIL$^{\star}$ \cite{ST-CIL} & DeepLab V3 & 71.4 & 40.0 & 63.6 & 76.7 & 54.3 & 71.1 & N/A & N/A & N/A & N/A & N/A & N/A\\
SSUL$^{\star}$ \cite{SSUL} & DeepLab V3 & 78.06 & 28.54 & 66.27 & 77.42 & 47.16 & 70.21 & 71.17 & 45.38 & 52.75 & 73.78 & 41.13 & 58.23\\
SSUL-M$^{\star}$ \cite{SSUL} & DeepLab V3 & 78.92 & 43.86 & 70.58 & 79.53 & 52.87 & 73.19 & 72.91 & 49.02 & 55.85 & \underline{74.79} & 48.87 & 62.45 \\
\midrule
Joint Training & SegFormer B2 & 80.84 & 74.97 & 79.44 & 80.84 & 74.97 & 79.44 & 78.36
& 79.87 & 79.44 & 80.46 & 78.32 & 79.44 \\
ILT \cite{ILT} & SegFormer B2 & 17.44 & 12.13 & 16.18 & 49.07 & 53.97 & 50.24 & 13.20 & 15.43 & 14.79 & 6.67 & 6.13 & 6.41 \\
MiB \cite{MiB} & SegFormer B2 & 73.21 & 37.93 & 64.81 & 78.78 & 60.93 & 74.53 & 61.12 & 58.02 & 58.56 & 48.7 & 39.58 & 44.36 \\
PLOP \cite{PLOP} & SegFormer B2 & 64.59 & 37.23 & 58.08 & 72.51 & 48.37 & 66.76 & 35.65 & 32.71 & 33.54 & 48.53 & 33.71 & 41.47 \\ 
SSUL \cite{SSUL} & SegFormer B2 & 79.91 & 40.56 & 70.54 & 79.91 & 56.83 & 74.41 & 74.33 & 60.79 & 64.66 & 74.06 & 51.85 & 63.48 \\
\midrule
ILT-M & SegFormer B2 & 15.15 & 11.01 & 14.16 & 49.85 & 53.49 & 50.72 & 12.91 & 15.43 & 14.71 & 6.75 & 6.07 & 6.42\\
MiB-M & SegFormer B2 & 73.89 & 58.39 & 70.72 & 79.91 & 63.56 & 75.20 & 70.11 & 65.17 & 66.58 & 69.73 & 56.28 & 63.33\\
PLOP-M & SegFormer B2 & 71.11 & 52.61 & 66.70 & 78.53 & \underline{65.58} & 75.44 & 67.29 & 62.91 & 64.16 & 57.94 & 51.64 & 54.94\\
SSUL-M & SegFormer B2 & \underline{79.84} & 49.33 & 72.58 & 79.84 & 55.82 & 74.12 & \textbf{76.04} & 61.95 & 65.98 & 74.23 & 52.24 & \underline{63.76} \\
\midrule
\textbf{SATS} (ours) & SegFormer B2 & 78.38 & \underline{62.02} & \underline{74.48} & \underline{80.24} & 61.17 & \underline{75.70} & 75.43 & \underline{64.13} & \underline{67.36} & 64.27 & \underline{58.66} & 61.60\\
\textbf{SATS-M} (ours) & SegFormer B2 & \textbf{80.37} & \textbf{64.54} & \textbf{76.61} & \textbf{81.44} & \textbf{70.02} & \textbf{78.72} & \underline{75.58} & \textbf{69.67} & \textbf{71.36} &  \textbf{76.21} & \textbf{61.62} & \textbf{69.27}\\
% 76.49 & 42.3 & 68.35
%  \midrule
\bottomrule
\end{tabular}
} %< \resizebox

\label{tab:voc}
\end{table*}

\subsection{Quantitative evaluation}

Multiple continual learning settings (15-1, 15-5, 5-3, 10-1 on Pascal VOC2012, and 100-10, 25-25 on ADE20K) {were} adopted to compare our SATS method with state-of-the-art methods for the two datasets, VOC2012 and ADE20K, {respectively. For} the VOC2012 dataset, among the prior methods for continual semantic segmentation with the DeepLab series (mainly DeepLab V3), SSUL performed the best regardless of  {memory} usage. However, the performance of SSUL on continually learned classes ({that is, except for} the classes learned in the first learning stage) was outperformed by other methods that used large amounts of auxiliary unlabeled data,  {such as} RECALL~\cite{RECALL}, with the setting VOC 15-1 (classes 16-20, 47.8\% vs. 43.86\%), and ST-CIL with VOC 15-5 (class 16-20, 54.3\% vs. 52.87\%). This  {phenomenon} is mainly because SSUL fixes the majority of the models and only updates the segmentation head during continual learning. {However, this} limits its ability  {to learn} new classes. Compared with the reported results (Table~\ref{tab:voc}, rows 2-10) of the strong baseline from their original work, where the DeepLab backbone was used,  {the proposed} SATS without using the memory of old data (Table~\ref{tab:voc}, second last row,  {“SATS”) outperformed} all existing methods without using memory by a significant margin and even  {outperformed} the state-of-the-art method (SSUL-M*), which used stored small old data during continual learning, except for the VOC 10-1 setting.

 {Using} the same SegFormer backbone for a fair comparison, our SATS method without memory (“SATS”) still outperformed the re-implemented representative methods ILT, MiB, PLOP, and SSUL. (Table~\ref{tab:voc}, rows 12-15), except for the VOC 10-1 setting, {in which the proposed} method was slightly worse than SSUL for the average classification of all classes (61.60\% vs. 63.48\%) but significantly better than SSUL for the learned new classes (classes 11–20, 58.66\% vs. 51.85\%) over the ten stages of continual learning. Again, this phenomenon could {be caused by SSUL's fixing of} the model parameters for the first set of 11 classes (ten foreground classes plus one background class) and only adds channels at the last layer for new classes.  {As a result,} SSUL often  {exhibits} superior performance on the initially learned old classes, but the model is too stable to learn new classes, particularly in more stages of learning. Compared with the freezing strategy in SSUL, our proposed method allows flexible updates of all model parameters and enables the updated segmentation model to easily learn new knowledge over continual learning.

\begin{figure*}[h]
    \centering
    \includegraphics[width=0.46\linewidth]{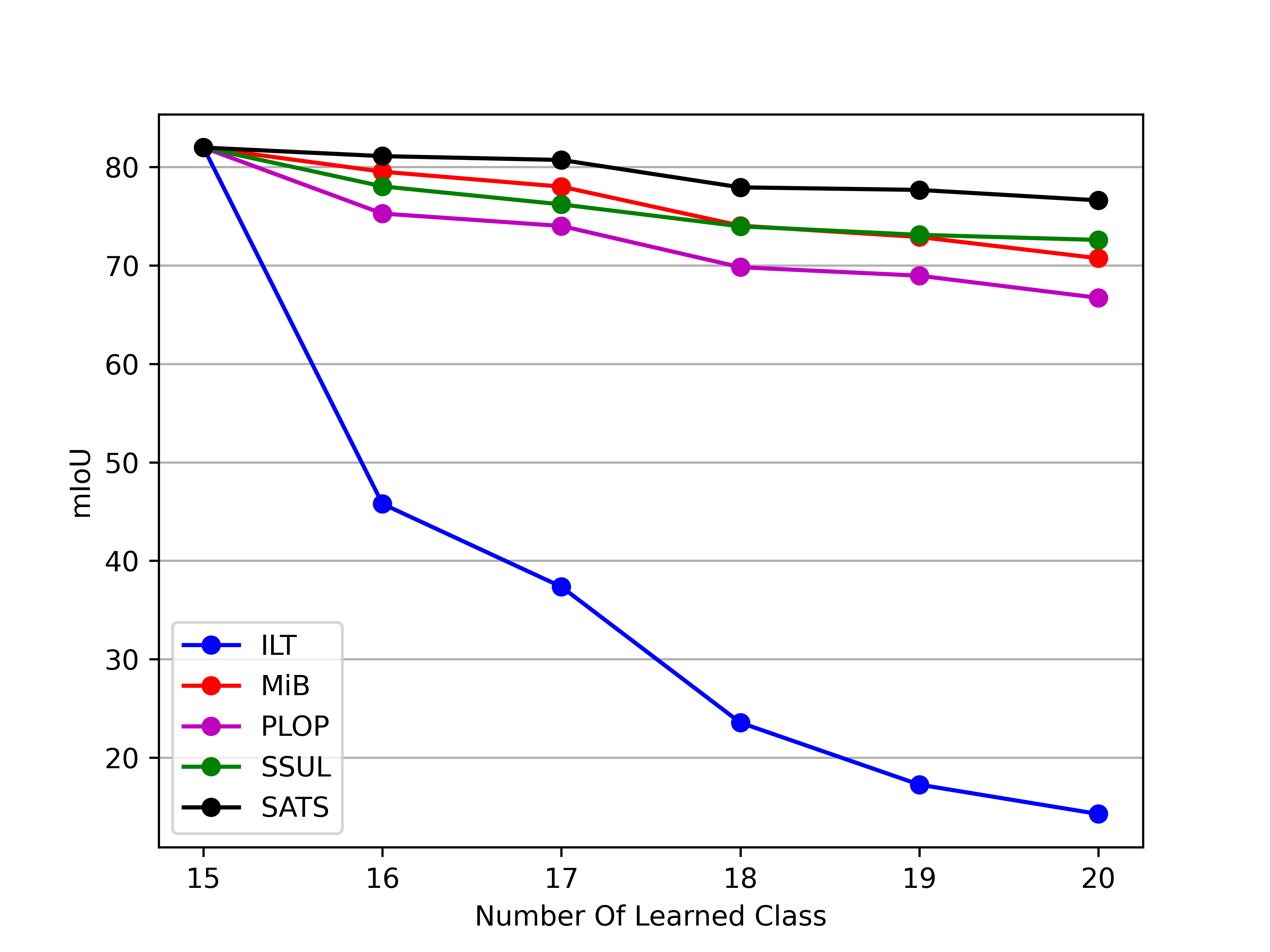}
 %   \hspace{\fill}
    \includegraphics[width=0.46\linewidth]{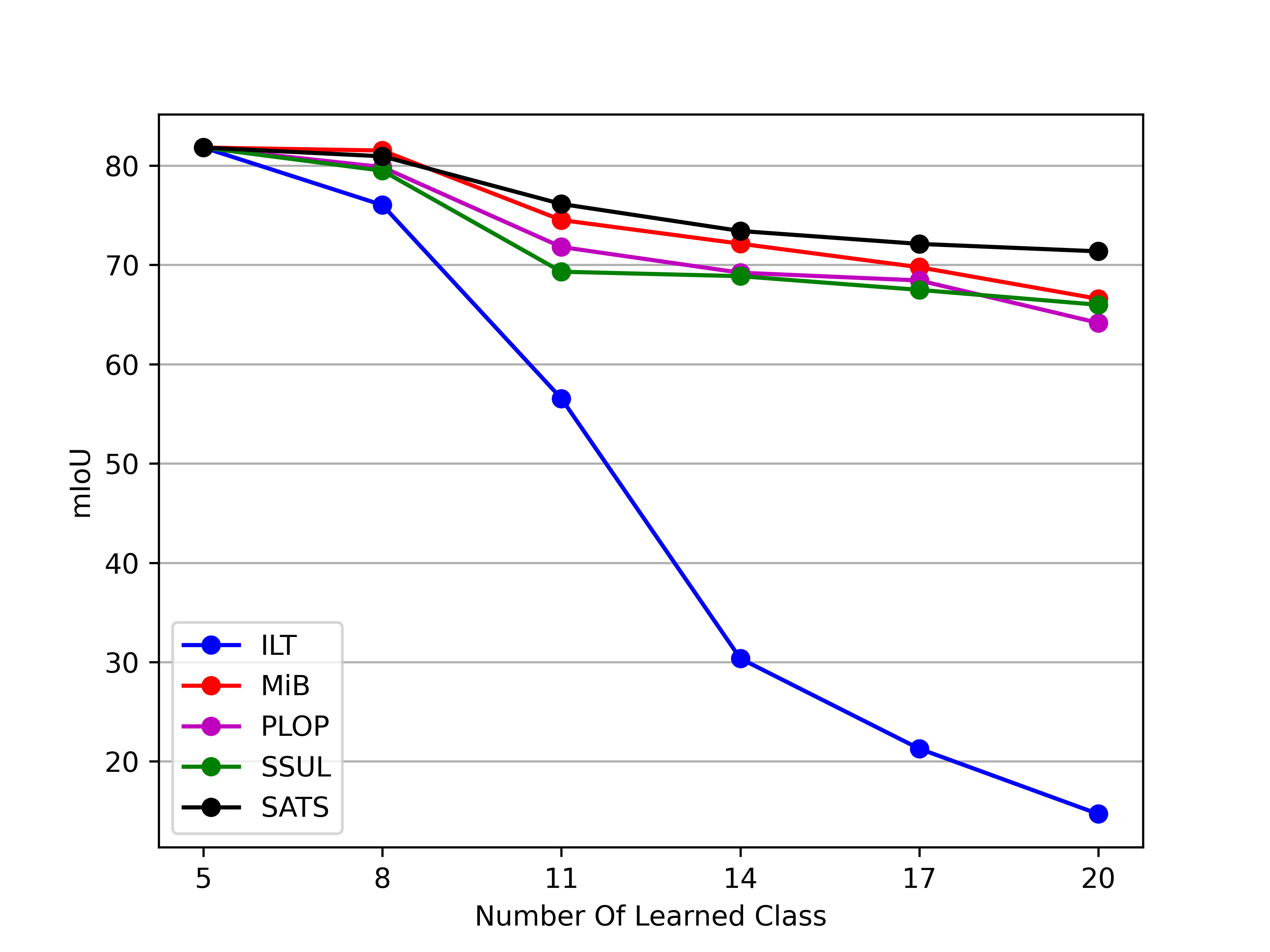}
    \caption{\normalsize Semantic segmentation performance at each stage of continual learning. Our SATS method always outperforms all existing methods after each stage of learning, with the protocol VOC 15-1 (left) and 5-3 (right). All methods used memory of the same size to store limited old data during continual learning.}
    \label{fig:effective}
\end{figure*}

This result is further confirmed by another series of experiments  {in which} all the methods used stored small old data during continual learning (Table~\ref{tab:voc}, rows 16-19 and the last row). With the help of stored old data, both our method (‘SATS-M’) and baselines MiB-M and PLOP-M substantially boosted performance. However, the improvement in the SSUL method (‘SSUL-M’) is limited, particularly for the settings VOC 15-5, 5-3, and 10-1.  {Because the} SSUL fixes model parameters for old classes, storing small amounts of old data would play a {limited} role in keeping old knowledge from forgetting.

In comparison, our proposed method allows the model to be updated flexibly for new class learning and simultaneously  makes {superior} use of stored old data to alleviate the forgetting of old knowledge. Consequently, our method (‘SATS-M’) achieves state-of-the-art performance in all four VOC settings for continual semantic segmentation. This result is further confirmed in Figures~\ref{fig:methodcompare} and~\ref{fig:effective},  {which reveal that the method outperformed the} representative methods at each stage of continual learning and the performance gap between existing methods and gradually  {increases} with more stages of learning.

% \newpage

Similar results were obtained  {for} the ADE20K dataset (see Table~\ref{tab:ADE}). For both settings, 100-10 and 25-25, notice that our method (last two rows) outperformed strong baselines on the new classes (columns 2 and 5) and all classes (columns 3 and 6),  {regardless of} whether memory buffer is used (rows 4-6 vs. row 8) (rows 1-3 vs. row 7). These results support {the generalizability of} our method to different continual semantic segmentation tasks.

\begin{table*}[t]
\caption{\normalsize Performance comparison on ADE20k dataset. }
\centering
\resizebox{0.6\linewidth}{!}
{ %< auto-adjusts font size to fill line
\begin{tabular}{ c|ccc|ccc } 
\toprule
\multirow{2}{*}{Method}& \multicolumn{3}{c|}{ADE 100-10 (6 Tasks)} & \multicolumn{3}{c}{ADE 25-25 (6 Tasks)}\\
& 0-100 & 101-150 & All & 0-25 & 26-150 & All \\ 
\midrule
MiB & 40.30 & 17.27 & 32.67 & 54.04 & 23.6 & 28.84\\
PLOP & 39.16 & 15.08 & 31.19 & 57.44 & 12.76 & 20.41 \\
SSUL & \underline{42.51} & 16.03 & 33.74 & \underline{59.43} & 13.87 & 21.88\\
\midrule
MiB-M & 41.12 & 18.86 & 33.75 & 54.59 & 24.25 & 29.47 \\
PLOP-M &  41.24 & 14.94 & 32.36 & 57.25 & 14.42 & 21.80\\
SSUL-M & \textbf{42.79} & 15.84 & 33.86 & \textbf{60.12} & 16.89 & 24.33\\
\midrule
\textbf{SATS} (ours) & 41.42 & \underline{19.09} & \underline{34.18} & 57.12 & \underline{26.23} & \underline{31.56}\\
\textbf{SATS-M} (ours) &  41.55 & \textbf{23.13} & \textbf{35.45} & 57.42 &\textbf{27.14}& \textbf{32.36}\\
\bottomrule
\end{tabular}
} %< \resizebox
\label{tab:ADE}
\end{table*}

\subsection{Visualization analysis}

Besides the quantitative evaluation, qualitative evaluation was performed with the VOC 15-1 protocol. A set of representative and challenging test input images were selected.  {In these images,} the foreground classes in these images were learned in the first (initial) stage of learning. The images  {were segmented after} each segmentation model completed all stages of continual learning. Therefore, if {old knowledge is catastrophically forgotten} over learning stages, the model would not  {effectively} segment these images containing earlier learned old classes.

\begin{figure*}[hbt]
    \centering
    \includegraphics[width=1.0\linewidth]{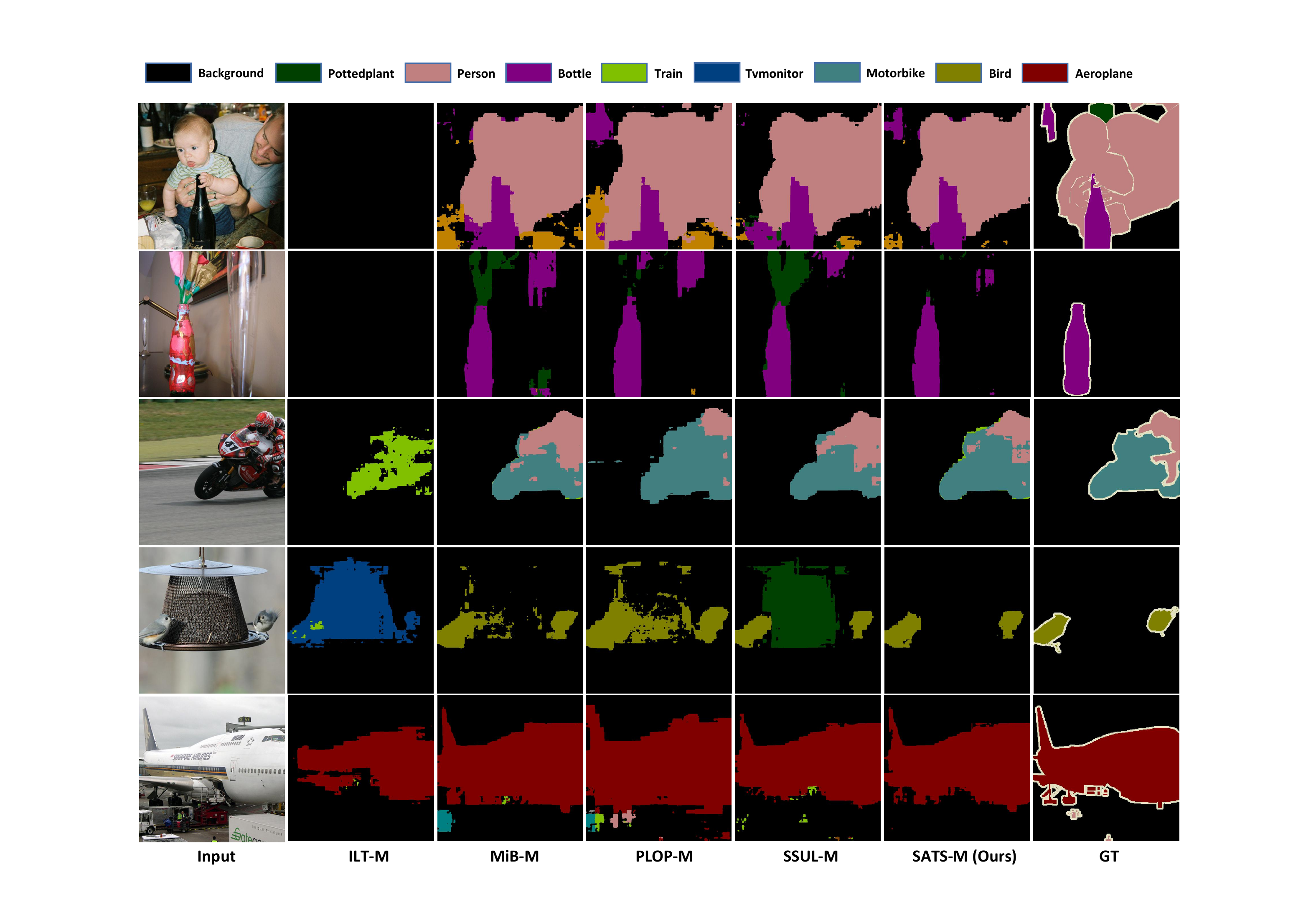}
    \caption{\normalsize Demonstration of segmentation results based on our SATS method and four representative baselines. }
    \label{fig:vis}
\end{figure*}

Figure~\ref{fig:vis}  {reveals that} all baselines except the ILT-M method can find the foreground regions. However, all of  {these} incorrectly segmented parts of the background regions (particularly those background regions close to the boundary of foreground objects) as foreground classes which {even did not} appear in the images. Our proposed method (second last column) can accurately discriminate the foreground regions against the background regions,  {regardless of} whether the foreground object is large (first and last rows) or small (fourth row). Another observation is that the proposed method can find more accurate boundaries of foreground objects particularly when multiple foreground objects appear in the images (first, third, and fourth  {rows, respectively). Superior} discrimination between background and foreground classes and between different foreground classes of regions  {using} our SATS method may partly  {originate} from the self-attention transfer, which can capture between-class relationships for knowledge distillation,  {whereas} all existing methods do not directly consider such relationships when distilling old knowledge.

% {
It's worth noting that this study focuses on alleviating the catastrophic forgetting of old knowledge in continual semantic segmentation rather than on the more accurate segmentation of challenging objects (e.g., very small objects). The missing “person” from the segmentation result (Figure~\ref{fig:vis}, last row, sixth column) is probably because the adopted SegFormer backbone may not be  {sufficiently} powerful for segmentation of very small objects in images. The other methods with SegFormer (Figure~\ref{fig:vis}, columns 2-5) {do} not  {accurately} segment the very small region of “person”. If a more powerful segmentation model can segment small objects, the proposed CRP would better extract the visual knowledge of small objects (from the corresponding class-specific region) and then transfer old knowledge to the new model during continual learning. Therefore,  {a more accurate segmentation of} small objects can be considered as one segmentation challenge, which is independent of the catastrophic forgetting issue in continual learning.
% }

\subsection{Ablation and sensitivity study}

The effect of each component of the proposed SATS method is confirmed  {through an} ablation study. In particular, when removing pseudo-labeling (Table~\ref{tab:ablation}, left, first row), the self-attention transfer loss $\mathcal{L}_a$ (left, second row), unbiased knowledge distillation loss $\mathcal{L}_d$ (left, third row), and model performance all decreased substantially (by 4\%–5\% in mIoU) for all three VOC settings (15-1, 5-3, 10-1) compared  {with} using all (the last row). This  {phenomenon} suggests that self-attention transfer may be used as the default component in future studies  {to improve} continual learning performance,  {which is similar to} the widely used knowledge distillation based on the model output and the proposed pseudo-labeling strategy.

 {In addition,} the necessity of using multiple scales of self-attention maps  {has been} confirmed. As  {displayed in} Table~\ref{tab:ablation} (right), when using self-attention information from only the last one, two, or three SegFormer blocks, the model performance  {degraded} substantially from a mIoU of 76.6\% to 71.5\%, 73.1\%, and 75.28\%. This result is reasonable because SegFormer collects visual features from all four scales of encoder blocks for accurate segmentation,  {indicating} that the self-attention information from the four encoder blocks may capture within- and between-class relationships at different visual scales, and  {such multiscale} information would represent more relational knowledge.

\begin{table*}[tbp]
    \caption{\normalsize Ablation study of the proposed SATS. Left: effect of SATS components on continual learning with three VOC settings. Right: the necessity of using multiscale self-attention transfer. PL: pseudo-labeling. }
    \centering
    \small{
        \setlength{\tabcolsep}{1mm}{
            \centering
            \begin{subtable}[hbtp]{0.55\linewidth}
                % \resizebox{\textwidth}{13mm}{
                \begin{tabular}{ c|c|c|ccc|ccc|ccc } 
                    \toprule
                    \multirow{2}{*}{PL} & \multirow{2}{*}{$L_a$} & \multirow{2}{*}{$L_d$} & \multicolumn{3}{c|}{VOC 15-1} & \multicolumn{3}{c|}{VOC 5-3} & \multicolumn{3}{c}{VOC 10-1}\\
                    &&& 0-15 & 16-20 & all & 0-5 & 6-20 & all & 0-10 & 11-20 & all \\ 
                    \midrule
                    
                    \usym{2613} & \checkmark & \checkmark  & 74.04 & 66.18 & 72.17 & 76.56 & 65.89 & 68.94 & 70.92 & 58.65 & 65.08\\ 
                    \checkmark & \usym{2613} & \checkmark  & 76.88 & 58.84 & 72.58 & 75.43 & 62.51 & 66.2 & 74.77 & 52.86 & 64.33	\\
                    \checkmark & \checkmark & \usym{2613}& 78.14 & 49.14 & 71.24 & 68.05 & 64.9 & 65.78 & 72.03 & 60.20 & 66.40\\
                    \checkmark & \checkmark & \checkmark & \textbf{80.37} & \textbf{64.54} & \textbf{76.61} & \textbf{75.58} & \textbf{69.67} & \textbf{71.36} &  \textbf{76.21} & \textbf{61.62} & \textbf{69.27} \\
                    \bottomrule
                \end{tabular}
                % }
            \end{subtable}
        }
    }
    \hspace{12mm}
    \small{
        \setlength{\tabcolsep}{1mm}{
            \begingroup
            \begin{subtable}[hbtp]{0.3\linewidth}
                \centering
                % \resizebox{\textwidth}{13mm}{ %< auto-adjusts font size to fill lin
                \begin{tabular}{ c|ccc }
                \toprule
                \multirow{2}{*}{Blocks for SATS} & \multicolumn{3}{c}{VOC 15-1} \\
                & 0-15 & 16-20 & all \\ 
                \midrule
                Last one block & 78.54 & 48.96 & 71.50 \\
                Last two blocks & 79.23 & 53.40 & 73.10 \\
                Last three blocks & 79.47 & 61.83 & 75.28 \\
                All blocks & \textbf{80.37} & \textbf{64.54} & \textbf{76.61}\\
                \bottomrule
                \end{tabular}
                % } %< \resizebox
            \end{subtable}
            \endgroup
        }
    }
    \label{tab:ablation}
\end{table*}

 {Furthermore, a} comparison with distillation using the encoder’s feature maps supports that the distillation of self-attention is more effective (Table~\ref{tab:ablation2}, row 1 vs. row 3), and the  {results of distilling both} is slightly worse than our results (Table~\ref{tab:ablation2}, row 2 vs. row 3). Furthermore, the ablation of CRP (Table~\ref{tab:ablation2}, rows 4-5 vs. row 6) shows that simply pooling all pixels for self-attention distillation (GP in Table~\ref{tab:ablation2}) or trivially distilling all self-attention vectors without pooling (NP in Table~\ref{tab:ablation2}) downgraded model performance. These results support the necessity of distillation of self-attention information and the  {performance improvement from} the region pooling operators.  
 
 {One additional} ablation study was performed by replacing the transformer backbone with the DeepLab V3 backbone while keeping the other {components (including CRP on feature map to perform knowledge distillation)} unchanged during continual semantic segmentation. Table~\ref{tab:ablation2} (row 7 vs. row 8) shows that the ablated version  {exhibits inferior performance to} the original version, supporting {the fact} that the relational knowledge from self-attention maps in the transformer backbone is crucial during continual semantic segmentation.

\begin{table*}[t]
    \caption{
    {\normalsize Ablation study of self-attention distillation, class-specific region pooling (CRP), and the model backbone. "Feature": encoder’s feature maps for distillation. GP: global pooling of self-attention over the whole image region. NP: no pooling.  }
   } % \caption
    \centering
    \small
    \resizebox{0.9\linewidth}{!}{
        \setlength{\tabcolsep}{1mm}{
        \setlength\extrarowheight{1.5pt}{
                \centering
                \begin{tabular}{ c|c|ccc|ccc } 
                    \toprule
                    \multicolumn{2}{c|}{\multirow{2}{*}{Setting}} & \multicolumn{3}{c|}{VOC 15-1 (6 Tasks)} & \multicolumn{3}{c}{VOC 10-1 (11 Tasks)}\\
                    \multicolumn{2}{c|}{} & 0-15 & 16-20 & All & 0-10 & 11-20 & all \\ 
                    \midrule
                    {\multirow{3}{*}{Distillation}} &Feature & 78.93 & 59.77 & 74.37 & 74.73 & 59.16 & 67.32 \\
                    & Self-Attention + Feature & \textbf{80.88} & 61.78 & 76.34 & \textbf{76.52} & 59.89 & 68.60\\
                    & Self-Attention (Ours) & 80.37 & \textbf{64.54} & \textbf{76.61} & 76.21 & \textbf{61.62} & \textbf{69.27}\\
                    \midrule
                    {\multirow{3}{*}{Pooling}}& GP & 80.32 & 61.67 & 75.88 & 74.38 & 60.81 & 67.92\\
                    & NP & 80.07 & 61.54 & 75.66 & 73.51 & 58.92 & 66.57\\
                    & CRP (Ours) & \textbf{80.37} & \textbf{64.54} & \textbf{76.61} & \textbf{76.21} & \textbf{61.62} & \textbf{69.27}\\
                    \midrule
                    {\multirow{2}{*}{ Backbone}} & CRP with DeepLab V3  & 78.00 & 45.65 & 70.3 & 66.61 & 44.43 & 56.05 \\
                    & CRP with Transformer (Ours) & \textbf{80.37} & \textbf{64.54} & \textbf{76.61} & \textbf{76.21} & \textbf{61.62} & \textbf{69.27} \\
                    \bottomrule
                \end{tabular}
        }}
        \hspace{\fill}
    } %< \resizebox

\label{tab:ablation2}
\end{table*}

\begin{figure}[h]
  \centering
%   \fbox{\rule{0pt}{0.5in} \rule{0.9\linewidth}{0pt}}
  \includegraphics[height=0.35\linewidth,width=0.45\linewidth]{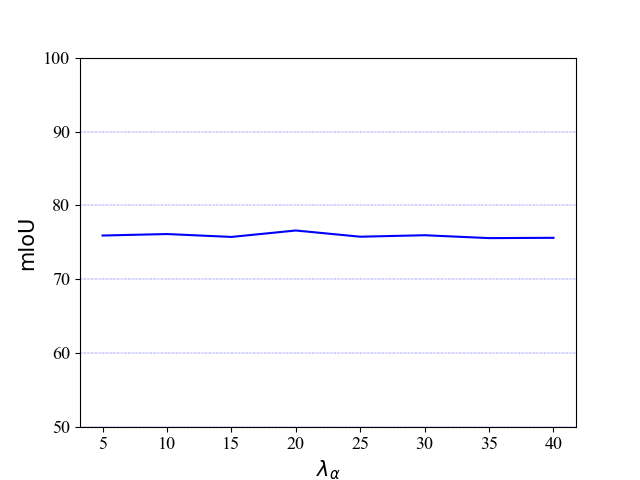}
  \includegraphics[height=0.35\linewidth,width=0.45\linewidth]{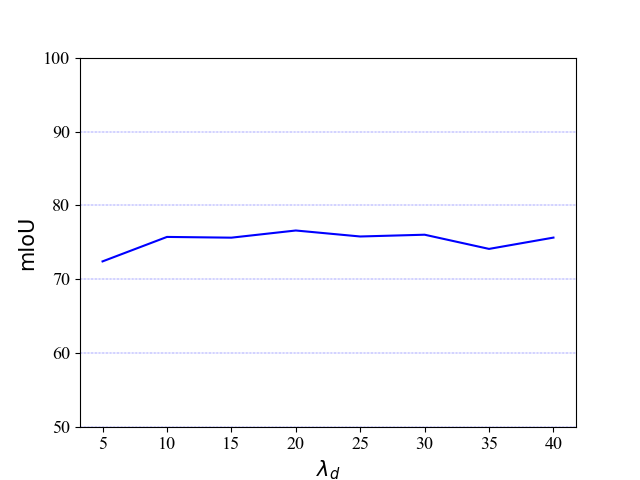}
  \vspace{-0.2cm}
    \caption{Sensitivity study of hyper-parameters  $\lambda_a$ and $\lambda_d$ with the continual learning setting VOC 15-1. When varying one hyper-parameter, the other one is fixed to 20. 
  }
  \label{fig:hyper-param-curve}
\end{figure}

% {
Finally, the sensitivity of the hyperparameters $\lambda_a$ and $\lambda_d$ in the loss function {is investigated.} As Figure~\ref{fig:hyper-param-curve} (left) shows that when $\lambda_a$ varies from 5 to 40, the model performance fluctuates within a small range. Similarly, varying $\lambda_d$ within the range $[10, 30]$ results in stable model performance (Figure~\ref{fig:hyper-param-curve}, right). This sensitivity study {reveals} that our method is robust to the setting of each hyperparameter within a large range of values.
% } 

% the sensitivity study of the hyper-parameters $\lambda_a$ and $\lambda_d$ in the loss function was performed. Figure~\ref{fig:hyper-param-curve} below shows that our method is robust to these hyper-parameters within a large range of values.

\subsection{Limitations}

Although extensive evaluations confirmed the effectiveness of the proposed SATS method on continual semantic segmentation, this study {has several limitations.} First, all experiments {were conducted according to} previous studies, where the maximum number of total learning stages {is} low.  It remains unclear 
how effective our method and other existing strategies are in more (e.g. 20 or 50) stages of continual learning. Second, the setting of the hyperparameter $\lambda_a$ for the self-attention transfer loss ${L}_a$ is affected by multiple factors. For example, without pseudo-labeling or storing a small amount of old data, the number of image regions belonging to old classes {decreases,} causing loss term ${L}_a$ much smaller (compared with the cross-entropy loss term, ${L}_c$). In this case, $\lambda_a$ needs to be set to a larger constant for the self-attention transfer loss ${L}_a$ to play its role during continual learning. A more adaptive setting of the hyperparameter $\lambda_a$ should be further explored. {Furthermore,} the proposed SATS method can be applied to other continual tasks including continual classification learning. {However, such an} extension has not been investigated in this study.

\section{Conclusion}\label{sec:conclusion}

In this study, a novel method, SATS, is proposed to effectively transfer within-class and between-class relational knowledge information from the old model to the new model during continual semantic segmentation. The relationship information can be {obtained using} self-attention maps from vision transformer models such as SegFormer, {which provides complementary knowledge not satisfactorily captured} by the output of the conventional convolution layers or fully connected layers in deep learning models. The proposed class-specific region pooling over self-attention maps can provide more efficient representations of both the within- and between-class knowledge in each image. {Such an efficient representation} allows the model to be flexibly updated. {Therefore, a superior trade-off exists} between model stability and flexibility during continual learning. The proposed SATS method is not only complementary to some knowledge distillation strategies but also to other types of continual semantic segmentation strategies including the pseudo-labeling strategy, and an example replay strategy. We believe that the proposed SATS can become a useful plug-and-play component that can be flexibly embedded in existing or future continual semantic segmentation frameworks. One by-product of this study is the observation of the superior performance of the transformer model over conventional convolutional models on continual semantic segmentation regardless of strategies used for continual learning. The proposed self-attention transfer strategy and the transformer-style models would also work  for other continual learning tasks, such as continual classification and detection learning. However, this study is limited in exploring the effectiveness of the proposed SATS method under more continual learning stages or continual learning without storing any old data, which will be investigated in future work.

\end{spacing}


% \nolinenumbers

\noindent \textbf{\\Acknowledgement}\\
This work is supported in part by the National Natural Science Foundation of China (grant No. 62071502) and the Guangdong Key Research and Development Program (grant No. 2020B1111190001).

%% References


% \bibliographystyle{elsarticle-num}

\bibliographystyle{IEEEtran}
% \bibliography{main_aaa}
\bibliography{sats_article}

\begin{thebibliography}{10}
\providecommand{\url}[1]{#1}
\csname url@samestyle\endcsname
\providecommand{\newblock}{\relax}
\providecommand{\bibinfo}[2]{#2}
\providecommand{\BIBentrySTDinterwordspacing}{\spaceskip=0pt\relax}
\providecommand{\BIBentryALTinterwordstretchfactor}{4}
\providecommand{\BIBentryALTinterwordspacing}{\spaceskip=\fontdimen2\font plus
\BIBentryALTinterwordstretchfactor\fontdimen3\font minus
  \fontdimen4\font\relax}
\providecommand{\BIBforeignlanguage}[2]{{%
\expandafter\ifx\csname l@#1\endcsname\relax
\typeout{** WARNING: IEEEtran.bst: No hyphenation pattern has been}%
\typeout{** loaded for the language `#1'. Using the pattern for}%
\typeout{** the default language instead.}%
\else
\language=\csname l@#1\endcsname
\fi
#2}}
\providecommand{\BIBdecl}{\relax}
\BIBdecl

\bibitem{LwF}
Z.~Li and D.~Hoiem, ``Learning without forgetting,'' \emph{IEEE Transactions on
  Pattern Analysis and Machine Intelligence}, vol.~40, no.~12, pp. 2935--2947,
  2018.

\bibitem{iCaRL}
S.-A. Rebuffi, A.~Kolesnikov, G.~Sperl, and C.~H. Lampert, ``i{C}a{RL}:
  Incremental classifier and representation learning,'' in \emph{Proceedings of
  the IEEE Conference on Computer Vision and Pattern Recognition}, 2017, pp.
  2001--2010.

\bibitem{UCIR}
S.~Hou, X.~Pan, C.~C. Loy, Z.~Wang, and D.~Lin, ``Learning a unified classifier
  incrementally via rebalancing,'' in \emph{Proceedings of the IEEE Conference
  on Computer Vision and Pattern Recognition}, 2019, pp. 831--839.

\bibitem{LwM}
P.~Dhar, R.~V. Singh, K.-C. Peng, Z.~Wu, and R.~Chellappa, ``Learning without
  memorizing,'' in \emph{Proceedings of the IEEE Conference on Computer Vision
  and Pattern Recognition}, 2019, pp. 5138--5146.

\bibitem{ILT}
U.~Michieli and P.~Zanuttigh, ``Incremental learning techniques for semantic
  segmentation,'' in \emph{Proceedings of the IEEE International Conference on
  Computer Vision Workshops}, 2019.

\bibitem{MiB}
F.~Cermelli, M.~Mancini, S.~R. Bulo, E.~Ricci, and B.~Caputo, ``Modeling the
  background for incremental learning in semantic segmentation,'' in
  \emph{Proceedings of the IEEE Conference on Computer Vision and Pattern
  Recognition}, 2020, pp. 9233--9242.

\bibitem{PLOP}
A.~Douillard, Y.~Chen, A.~Dapogny, and M.~Cord, ``{PLOP}: Learning without
  forgetting for continual semantic segmentation,'' in \emph{Proceedings of the
  IEEE Conference on Computer Vision and Pattern Recognition}, 2021, pp.
  4040--4050.

\bibitem{SDR}
U.~Michieli and P.~Zanuttigh, ``Continual semantic segmentation via
  repulsion-attraction of sparse and disentangled latent representations,'' in
  \emph{Proceedings of the IEEE Conference on Computer Vision and Pattern
  Recognition}, 2021, pp. 1114--1124.

\bibitem{CatastrophicForgetting}
M.~McCloskey and N.~J. Cohen, ``Catastrophic interference in connectionist
  networks: The sequential learning problem,'' in \emph{Psychology of Learning
  and Motivation}, vol.~24, 1989, pp. 109--165.

\bibitem{PLOPLong}
A.~Douillard, Y.~Chen, A.~Dapogny, and M.~Cord, ``Tackling catastrophic
  forgetting and background shift in continual semantic segmentation,''
  \emph{arXiv preprint arXiv:2106.15287}.

\bibitem{ST-CIL}
L.~Yu, X.~Liu, and J.~van~de Weijer, ``Self-training for class-incremental
  semantic segmentation,'' \emph{IEEE Transactions on Neural Networks and
  Learning Systems}, vol.~PP, 2022.

\bibitem{SSUL}
S.~Cha, Y.~Yoo, T.~Moon \emph{et~al.}, ``{SSUL}: Semantic segmentation with
  unknown label for exemplar-based class-incremental learning,'' in
  \emph{Advances in Neural Information Processing Systems}, vol.~34, 2021.

\bibitem{VOC}
M.~Everingham, L.~Van~Gool, C.~K. Williams, J.~Winn, and A.~Zisserman, ``The
  pascal visual object classes (voc) challenge,'' \emph{International Journal
  of Computer Vision}, vol.~88, no.~2, pp. 303--338, 2010.

\bibitem{ADE20K}
B.~Zhou, H.~Zhao, X.~Puig, S.~Fidler, A.~Barriuso, and A.~Torralba, ``Scene
  parsing through ade20k dataset,'' in \emph{Proceedings of the IEEE Conference
  on Computer Vision and Pattern Recognition}, 2017, pp. 633--641.

\bibitem{FCN}
J.~Long, E.~Shelhamer, and T.~Darrell, ``Fully convolutional networks for
  semantic segmentation,'' in \emph{Proceedings of the IEEE Conference on
  Computer Vision and Pattern Recognition}, 2015, pp. 3431--3440.

\bibitem{UNet}
O.~Ronneberger, P.~Fischer, and T.~Brox, ``U-{N}et: Convolutional networks for
  biomedical image segmentation,'' in \emph{International Conference on Medical
  Image Computing and Computer-Assisted Intervention}, 2015, pp. 234--241.

\bibitem{DeeplabV1}
L.-C. Chen, G.~Papandreou, I.~Kokkinos, K.~Murphy, and A.~L. Yuille, ``Semantic
  image segmentation with deep convolutional nets and fully connected crfs,''
  \emph{arXiv preprint arXiv:1412.7062}.

\bibitem{DeeplabV2}
------, ``Deeplab: Semantic image segmentation with deep convolutional nets,
  atrous convolution, and fully connected crfs,'' \emph{IEEE Transactions on
  Pattern Analysis and Machine Intelligence}, vol.~40, no.~4, pp. 834--848,
  2018.

\bibitem{DeeplabV3}
L.-C. Chen, G.~Papandreou, F.~Schroff, and H.~Adam, ``Rethinking atrous
  convolution for semantic image segmentation,'' \emph{arXiv preprint
  arXiv:1706.05587}.

\bibitem{DeeplabV3+}
L.-C. Chen, Y.~Zhu, G.~Papandreou, F.~Schroff, and H.~Adam, ``Encoder-decoder
  with atrous separable convolution for semantic image segmentation,'' in
  \emph{Proceedings of the European Conference on Computer Vision}, 2018, pp.
  801--818.

\bibitem{SETR}
S.~Zheng, J.~Lu, H.~Zhao, X.~Zhu, Z.~Luo, Y.~Wang, Y.~Fu, J.~Feng, T.~Xiang,
  P.~H. Torr, and L.~Zhang, ``Rethinking semantic segmentation from a
  sequence-to-sequence perspective with transformers,'' in \emph{Proceedings of
  the IEEE Conference on Computer Vision and Pattern Recognition}, 2021, pp.
  6881--6890.

\bibitem{SegFormer}
E.~Xie, W.~Wang, Z.~Yu, A.~Anandkumar, J.~M. Alvarez, and P.~Luo, ``Segformer:
  Simple and efficient design for semantic segmentation with transformers,'' in
  \emph{Advances in Neural Information Processing Systems}, vol.~34, 2021.

\bibitem{MaskFormer}
B.~Cheng, A.~G. Schwing, and A.~Kirillov, ``Per-pixel classification is not all
  you need for semantic segmentation,'' in \emph{Advances in Neural Information
  Processing Systems}, 2021.

\bibitem{SSSS_1}
Y.~Wang, H.~Wang, Y.~Shen, J.~Fei, W.~Li, G.~Jin, L.~Wu, R.~Zhao, and X.~Le,
  ``Semi-supervised semantic segmentation using unreliable pseudo-labels,'' in
  \emph{Proceedings of the IEEE Conference on Computer Vision and Pattern
  Recognition}, 2022, pp. 4238--4247.

\bibitem{SSSS_2}
L.~Yang, W.~Zhuo, L.~Qi, Y.~Shi, and Y.~Gao, ``St++: Make self-training work
  better for semi-supervised semantic segmentation,'' in \emph{Proceedings of
  the IEEE Conference on Computer Vision and Pattern Recognition}, 2022, pp.
  4258--4267.

\bibitem{SSSS_3}
J.~Peng, G.~Estrada, M.~Pedersoli, and C.~Desrosiers, ``Deep co-training for
  semi-supervised image segmentation,'' \emph{Pattern Recognition}, vol. 107,
  p. 107269, 2020.

\bibitem{WSSS_point}
A.~Bearman, O.~Russakovsky, V.~Ferrari, and L.~Fei-Fei, ``What's the point:
  Semantic segmentation with point supervision,'' \emph{arXiv preprint,
  arXiv:1506.02106}, 2015.

\bibitem{WSSS_Point2}
R.~A. McEver and B.~S. Manjunath, ``Pcams: Weakly supervised semantic
  segmentation using point supervision,'' \emph{arXiv preprint
  arXiv:2007.05615}, 2020.

\bibitem{WSSS_scribbles2}
Z.~Pan, P.~Jiang, Y.~Wang, C.~Tu, and A.~G. Cohn, ``Scribble-supervised
  semantic segmentation by uncertainty reduction on neural representation and
  self-supervision on neural eigenspace,'' in \emph{Proceedings of the IEEE
  International Conference on Computer Vision}, 2021, pp. 7396--7405.

\bibitem{WSSS_scribbles}
D.~Lin, J.~Dai, J.~Jia, K.~He, and J.~Sun, ``Scribblesup: Scribble-supervised
  convolutional networks for semantic segmentation,'' in \emph{IEEE Conference
  on Computer Vision and Pattern Recognition}, 2016, pp. 3159--3167.

\bibitem{Lee2021BBAMBB}
J.~Lee, J.~Yi, C.~Shin, and S.~Yoon, ``Bbam: Bounding box attribution map for
  weakly supervised semantic and instance segmentation,'' \emph{IEEE Conference
  on Computer Vision and Pattern Recognition}, pp. 2643--2651, 2021.

\bibitem{WSSS_bbox2}
J.~Dai, K.~He, and J.~Sun, ``Boxsup: Exploiting bounding boxes to supervise
  convolutional networks for semantic segmentation,'' in \emph{IEEE
  International Conference on Computer Vision}, 2015, pp. 1635--1643.

\bibitem{WSSS_bbox1}
T.~Zheng, Q.~Wang, Y.~Shen, X.~Ma, and X.~Lin, ``High-resolution rectified
  gradient-based visual explanations for weakly supervised segmentation,''
  \emph{Pattern Recognition}, vol. 129, p. 108724, 2022.

\bibitem{WSSS_RRM}
B.~Zhang, J.~Xiao, Y.~Wei, K.~Huang, S.~Luo, and Y.~Zhao, ``End-to-end weakly
  supervised semantic segmentation with reliable region mining,'' \emph{Pattern
  Recognition}, vol. 128, p. 108663, 2022.

\bibitem{WSSS_L2G}
P.~Jiang, Y.~Yang, Q.~Hou, and Y.~Wei, ``{L2G:} {A} simple local-to-global
  knowledge transfer framework for weakly supervised semantic segmentation,''
  in \emph{Proceedings of the IEEE Conference on Computer Vision and Pattern
  Recognition}, 2022, pp. 16\,865--16\,875.

\bibitem{WSSS_OAA}
P.-T. Jiang, L.-H. Han, Q.~Hou, M.-M. Cheng, and Y.~Wei, ``Online attention
  accumulation for weakly supervised semantic segmentation,'' \emph{IEEE
  Transactions on Pattern Analysis and Machine Intelligence}, vol.~44, no.~10,
  pp. 7062--7077, 2022.

\bibitem{WSSS_ref5}
S.~Yi, H.~Ma, X.~Wang, T.~Hu, X.~Li, and Y.~Wang, ``Weakly-supervised semantic
  segmentation with superpixel guided local and global consistency,''
  \emph{Pattern Recognition}, vol. 124, p. 108504, 2022.

\bibitem{WSSS_ref6}
S.~Kho, P.~Lee, W.~Lee, M.~Ki, and H.~Byun, ``Exploiting shape cues for weakly
  supervised semantic segmentation,'' \emph{Pattern Recognition}, vol. 132, p.
  108953, 2022.

\bibitem{IS_1}
Z.~Lin, Z.~Zhang, L.-Z. Chen, M.-M. Cheng, and S.-P. Lu, ``Interactive image
  segmentation with first click attention,'' in \emph{Proceedings of the IEEE
  Conference on Computer Vision and Pattern Recognition}, 2020, pp.
  13\,336--13\,345.

\bibitem{IS_2}
X.~Chen, Z.~Zhao, Y.~Zhang, M.~Duan, D.~Qi, and H.~Zhao, ``Focalclick: Towards
  practical interactive image segmentation,'' in \emph{Proceedings of the IEEE
  Conference on Computer Vision and Pattern Recognition}, June 2022, pp.
  1300--1309.

\bibitem{IS_3_MIDeepSeg}
X.~Luo, G.~Wang, T.~Song, J.~Zhang, M.~Aertsen, J.~Deprest, S.~Ourselin,
  T.~Vercauteren, and S.~Zhang, ``Mideepseg: Minimally interactive segmentation
  of unseen objects from medical images using deep learning,'' \emph{Medical
  Image Analysis}, vol.~72, p. 102102, 2021.

\bibitem{Chen2019DomainAF}
M.~Chen, H.~Xue, and D.~Cai, ``Domain adaptation for semantic segmentation with
  maximum squares loss,'' in \emph{Proceedings of the IEEE International
  Conference on Computer Vision}, 2019, pp. 2090--2099.

\bibitem{You_ref51}
Z.~Zheng and Y.~Yang, ``Rectifying pseudo label learning via uncertainty
  estimation for domain adaptive semantic segmentation,'' \emph{International
  Journal of Computer Vision}, vol. 129, no.~4, p. 1106–1120, 2021.

\bibitem{You_MM2021}
F.~You, J.~Li, L.~Zhu, Z.~Chen, and Z.~Huang, ``Domain adaptive semantic
  segmentation without source data,'' in \emph{Proceedings of the 29th ACM
  International Conference on Multimedia}, ser. MM '21, 2021, p. 3293–3302.

\bibitem{You_arxiv2021}
F.~You, J.~Li, and Z.~Zhao, ``Test-time batch statistics calibration for
  covariate shift,'' \emph{arXiv preprint arXiv:2110.04065}, 2021.

\bibitem{You_MM2022}
F.~You, J.~Li, Z.~Chen, and L.~Zhu, ``Pixel exclusion: Uncertainty-aware
  boundary discovery for active cross-domain semantic segmentation,'' in
  \emph{Proceedings of the 30th ACM International Conference on Multimedia},
  ser. MM '22, 2022, p. 1866–1874.

\bibitem{End2End-IL}
F.~M. Castro, M.~J. Marin-Jimenez, N.~Guil, C.~Schmid, and K.~Alahari,
  ``End-to-end incremental learning,'' in \emph{Proceedings of the European
  Conference on Computer Vision}, 2018, pp. 233--248.

\bibitem{LL-PRDR}
S.~Hou, X.~Pan, C.~C. Loy, Z.~Wang, and D.~Lin, ``Lifelong learning via
  progressive distillation and retrospection,'' in \emph{Proceedings of the
  European Conference on Computer Vision}, 2018, pp. 437--452.

\bibitem{PODNet}
A.~Douillard, M.~Cord, C.~Ollion, T.~Robert, and E.~Valle, ``{PODN}et: Pooled
  outputs distillation for small-tasks incremental learning,'' in
  \emph{Proceedings of the European Conference on Computer Vision}, 2020, pp.
  86--102.

\bibitem{IL2M}
E.~Belouadah and A.~Popescu, ``Il2m: Class incremental learning with dual
  memory,'' in \emph{Proceedings of the IEEE International Conference on
  Computer Vision}, 2019, pp. 583--592.

\bibitem{MCS_rehearsal_continual}
C.~Zhuang, S.~Huang, G.~Cheng, and J.~Ning, ``Multi-criteria selection of
  rehearsal samples for continual learning,'' \emph{Pattern Recognition}, vol.
  132, p. 108907, 2022.

\bibitem{ES-CIL}
S.~Mittal, S.~Galesso, and T.~Brox, ``Essentials for class incremental
  learning,'' in \emph{IEEE Conference on Computer Vision and Pattern
  Recognition Workshops}, 2021, pp. 3508--3517.

\bibitem{AAN-CIL}
Y.~Liu, B.~Schiele, and Q.~Sun, ``Adaptive aggregation networks for
  class-incremental learning,'' in \emph{Proceedings of the IEEE Conference on
  Computer Vision and Pattern Recognition}, 2021, pp. 2544--2553.

\bibitem{PEKCL-TRANS}
Z.~Li, C.~Zhong, S.~Liu, R.~Wang, and W.-S. Zheng, ``Preserving earlier
  knowledge in continual learning with the help of all previous feature
  extractors,'' \emph{arXiv preprint arXiv:2104.13614}, 2021.

\bibitem{DFCIL}
J.~Smith, Y.-C. Hsu, J.~Balloch, Y.~Shen, H.~Jin, and Z.~Kira, ``Always be
  dreaming: A new approach for data-free class-incremental learning,'' in
  \emph{Proceedings of the IEEE International Conference on Computer Vision},
  2021, pp. 9374--9384.

\bibitem{RECALL}
A.~Maracani, U.~Michieli, M.~Toldo, and P.~Zanuttigh, ``{RECALL}: Replay-based
  continual learning in semantic segmentation,'' in \emph{Proceedings of the
  IEEE International Conference on Computer Vision}, 2021, pp. 7026--7035.

\bibitem{EWC}
J.~Kirkpatrick, R.~Pascanu, N.~Rabinowitz, J.~Veness, G.~Desjardins, A.~A.
  Rusu, K.~Milan, J.~Quan, T.~Ramalho, A.~Grabska-Barwinska, D.~Hassabis,
  C.~Clopath, D.~Kumaran, and R.~Hadsell, ``Overcoming catastrophic forgetting
  in neural networks,'' \emph{Proceedings of the National Academy of Sciences},
  vol. 114, no.~13, pp. 3521--3526, 2017.

\bibitem{CIL-DMC}
J.~Zhang, J.~Zhang, S.~Ghosh, D.~Li, S.~Tasci, L.~Heck, H.~Zhang, and C.-C.~J.
  Kuo, ``Class-incremental learning via deep model consolidation,'' in
  \emph{Proceedings of the IEEE Winter Conference on Applications of Computer
  Vision}, 2020, pp. 1131--1140.

\bibitem{DER}
S.~Yan, J.~Xie, and X.~He, ``{DER}: Dynamically expandable representation for
  class incremental learning,'' in \emph{Proceedings of the IEEE Conference on
  Computer Vision and Pattern Recognition}, 2021, pp. 3014--3023.

\bibitem{DER_Bayesian}
Y.~Yang, B.~Chen, and H.~Liu, ``Bayesian compression for dynamically expandable
  networks,'' \emph{Pattern Recognition}, vol. 122, p. 108260, 2022.

\bibitem{PVT}
W.~Wang, E.~Xie, X.~Li, D.-P. Fan, K.~Song, D.~Liang, T.~Lu, P.~Luo, and
  L.~Shao, ``Pyramid vision transformer: A versatile backbone for dense
  prediction without convolutions,'' in \emph{Proceedings of the IEEE
  International Conference on Computer Vision}, 2021, pp. 548--558.

\bibitem{SwinT}
Z.~Liu, Y.~Lin, Y.~Cao, H.~Hu, Y.~Wei, Z.~Zhang, S.~Lin, and B.~Guo, ``Swin
  transformer: Hierarchical vision transformer using shifted windows,'' in
  \emph{Proceedings of the IEEE International Conference on Computer Vision},
  2021, pp. 10\,012--10\,022.

\bibitem{Mix-ViT}
X.~Yu, J.~Wang, Y.~Zhao, and Y.~Gao, ``Mix-vit: Mixing attentive vision
  transformer for ultra-fine-grained visual categorization,'' \emph{Pattern
  Recognition}, vol. 135, p. 109131, 2023.

\bibitem{ACFNet}
F.~Zhang, Y.~Chen, Z.~Li, Z.~Hong, J.~Liu, F.~Ma, J.~Han, and E.~Ding,
  ``Acfnet: Attentional class feature network for semantic segmentation,'' in
  \emph{Proceedings of the IEEE International Conference on Computer Vision},
  2019, p. 6798{\textendash}6807.

\bibitem{CRP_1}
X.~He, J.~Liu, J.~Fu, X.~Zhu, J.~Wang, and H.~Lu, ``Consistent-separable
  feature representation for semantic segmentation,'' in \emph{{AAAI}
  Conference on Artificial Intelligence}, 2021, pp. 1531--1539.

\bibitem{CRP_2}
H.~Ma, X.~Lin, Z.~Wu, and Y.~Yu, ``Coarse-to-fine domain adaptive semantic
  segmentation with photometric alignment and category-center regularization,''
  in \emph{IEEE Conference on Computer Vision and Pattern Recognition}, 2021,
  pp. 4050--4059.

\bibitem{ImageNet}
J.~Deng, W.~Dong, R.~Socher, L.-J. Li, K.~Li, and L.~Fei-Fei, ``Image{N}et: A
  large-scale hierarchical image database,'' in \emph{IEEE Conference on
  Computer Vision and Pattern Recognition}, 2009, pp. 248--255.

\end{thebibliography}


% \bibliographystyle{IEEEtran}
% \bibliography{egbib}
% 
% \end{thebibliography}

\end{document}